\documentclass{article}

\usepackage[final]{corl_2020} % Uncomment for the camera-ready ``final'' version.

\usepackage{graphics} % for pdf, bitmapped graphics files
\usepackage{epsfig} % for postscript graphics files
\usepackage{amsmath} % assumes amsmath package installed
\usepackage[dvipsnames]{xcolor}
\usepackage{algorithm}
\usepackage{algorithmic}
\usepackage{booktabs}
\usepackage{subcaption}
\usepackage{hyperref}
\usepackage{authblk}

\definecolor{brightturquoise}{rgb}{0.03, 0.91, 0.87}

\usepackage{multirow}
\usepackage{array}

\usepackage{wrapfig}
\usepackage{threeparttable}
\usepackage{tikz}
\usetikzlibrary{bayesnet}
\usetikzlibrary{positioning}

\title{Learning a Decision Module \\
by Imitating Driver's Control Behaviors
}

\author{
  Junning Huang$^{1,*}$ Sirui Xie$^{2,*}$ Jiankai Sun$^{5,}$\thanks{indicates equal contribution.}~ Qiurui Ma$^{4}$ Chunxiao Liu$^{3}$ Dahua Lin$^{5}$ Bolei Zhou$^{5}$ \\
  $^1$Technische Universität Darmstadt $^2$University of California, Los Angeles $^3$SenseTime Research $^4$Hong Kong University of Science and Technology $^5$The Chinese University of Hong Kong  
}
\begin{document}
\maketitle

%===============================================================================

\begin{abstract}
  Autonomous driving systems have a pipeline of perception, decision, planning, and control. The decision module processes information from the perception module and directs the execution of downstream planning and control modules. On the other hand, the recent success of deep learning suggests that this pipeline could be replaced by end-to-end neural control policies, however, safety cannot be well guaranteed for the data-driven neural networks. In this work, we propose a hybrid framework to learn neural decisions  in the classical modular pipeline through end-to-end imitation learning.  This hybrid framework can preserve the merits of the classical pipeline such as the strict enforcement of physical and logical constraints while learning complex driving decisions from data. To circumvent the ambiguous annotation of human driving decisions, our method learns high-level driving decisions by imitating low-level control behaviors. We show in the simulation experiments that our modular driving agent can generalize its driving decision and control to various complex scenarios where the rule-based programs fail. It can also generate smoother and safer driving trajectories than end-to-end neural policies. Demo and code are available at \url{https://decisionforce.github.io/modulardecision/}.
\end{abstract}

\vspace{-5pt}
\section{Introduction}
\vspace{-5pt}
An autonomous driving system is essentially a decision-making system that takes a stream of onboard sensory data as input, processes it with prior knowledge about driving scenarios, and then make a reasonable \textit{decision}, and outputs \textit{control} signals to steer the vehicle \cite{paden2016survey}. Such a system is often modularized into perception, decision, planning, and control.  While the perception and its auxiliary, map reconstruction, tend to be more open to machine learning methods \cite{thrun2005probabilistic}, the downstream modules such as \textit{decision}, \textit{planning} and \textit{control} remain as program-based to ensure the safe interaction with the physical world. In other words, this safety guarantee is built upon a human-inspectable and interruptible basis.
% , such as \textcolor{blue}{we need a non-trivial example here}
However, in complex driving environments, it is extremely difficult to build a complete rule-based decision-making system. %As a matter of fact, the pursuit of completeness is always not iterative. 
So many corner cases are out there such that a significant program refactorization is required when one corner case is inconsistent with the existing rules. Towards a more flexible alternative, we explore a learning-based decision-making module whose downstream modules, i.e. planning and control, remain encapsulated. Essentially, we draw inspiration from a prior work \cite{perkins2002lyapunov}, which, in stark contrast to end-to-end control policies, learns a decision module to switch between controllers designed with Lyapunov domain knowledge.
% , combining the flexibility of data-driven decisions and the transparency of planning and control. 

\begin{figure*}[htp]
  \centering
  \vspace{-20pt}
  \includegraphics[width=\textwidth]{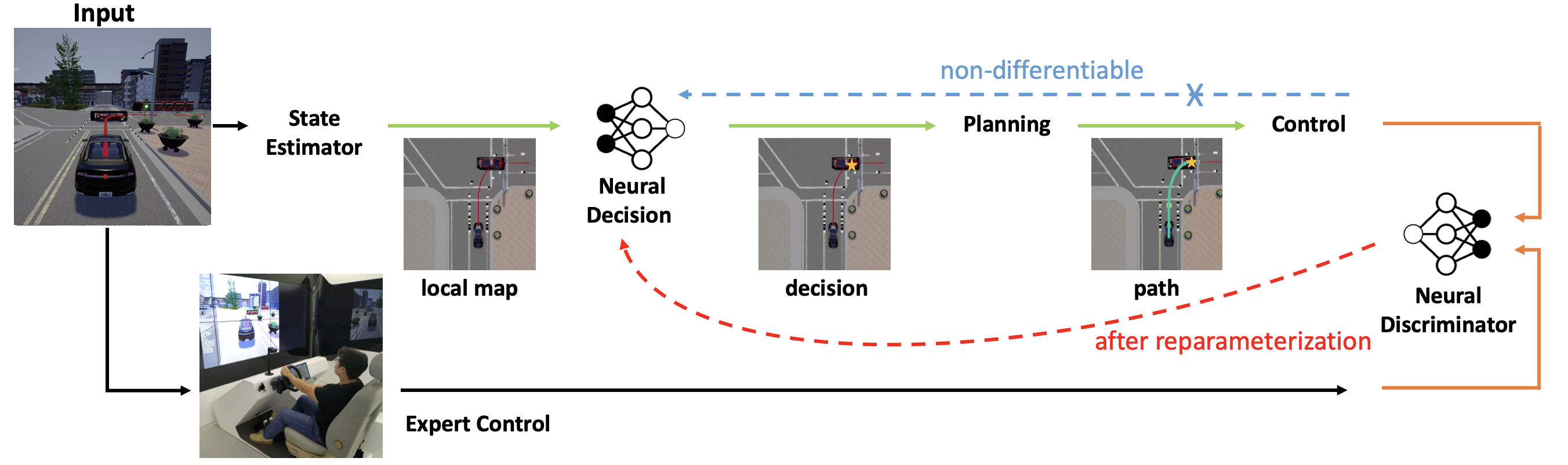}
  \caption{\textbf{Modular decision framework.} The green line shows the modular pipeline of the autonomous driving system.
    %, which is also the generation process of our framework. 
    Solid orange lines indicate the offline training of the neural discriminator, while the dashed blue line indicates that the planning and control modules are not differentiable. But the decision policy can be trained with reward assigned on control action by the discriminator after reparameterization, as indicated by the dashed red line. }
  \label{fig:overview}
  \vspace{-15pt}
\end{figure*}

The primary challenge we have for a learning-based decision module is the source of supervision. \textit{Driving  decision} is defined as the high-level abstraction about what lane the driver wants the vehicle to be with at which velocity in $T$ seconds. For an autonomous driving system, this decision determines the execution of downstream planning and control modules. For a human driver, however, it is both difficult and ambiguous to describe such a driving decision. Imagine two drivers try to merge into traffic at one roundabout, they may have different habitual behaviors to execute. Even the same individual may make different decisions in almost identical scenarios. To this end, there is hardly a set of golden criteria to evaluate the quality of human-annotated driving decisions. %, which would imply noisy labels for large-scale supervised learning. 
%The dilemma here is, if we want to evaluate labels directly at the decision level, a standardized set of criteria is inevitable. But its complexity is asymptotically no less than composing an FSM decision module. 
In contrast, drivers' physical behaviors, such as stepping on throttle/brake and steering the wheel, implicitly reflect their high-level decisions. Even though it is only a weak supervision in the sense that the decision is not directly supervised, it is easily accessible and technically reliable.

The setup of weak and indirect supervision requires a mechanism to infer human intentions from
their behaviors and imitate these cognitive decisions. Intention understanding and behavior imitation have been extensively studied in the field of artificial intelligence  and cognitive science, under the banner of Inverse Planning \cite{baker2009action} and Inverse Reinforcement Learning \cite{abbeel2004apprenticeship,9361118,9197185}. The former one mainly focuses on inferring the unobservable intention from the observation of human behaviors in a third-person view \cite{wang2013probabilistic}; The latter one adopts end-to-end neural policies and does not distinguish intentions from actions \cite{ross2011reduction,ziebart2008maximum}. Henceforth, neither of them discussed the formulation of indirect supervision as in this work. More specifically, our explicit representation of intentions and modular pipeline is appealing to physical systems like autonomous driving since the end-to-end neural counterparts do not guarantee the fulfillment of low-level constraints in states or actions \cite{fisac2018general}. Nor do they have robustness or stability guarantees \cite{huang2017adversarial}, which are of great importance to the real-world deployment of classical controllers. Given human driving behaviors, the expected framework should learn a high-level decision module in the hierarchical pipeline where downstream modules inherit the merits of the classical paradigm. Due to the non-differentiability of these downstream programs, a tailored learning scheme is needed.

In this work, we propose an imitation learning framework for a modular driving pipeline that \textit{learns neural decisions from human behavioral demonstrations}. The learning is conducted in a generative-adversarial manner \cite{goodfellow2014generative}. The generator (green line in Fig.\ref{fig:overview}) simulates the modularized driving pipeline. In the upstream, a neural decision module generates decisions according to the information from a local map. Decisions are then passed into the programmed downstream modules for planning and control. The generation process ends at the control module, where the interaction with the environment is triggered. For adversarial training (orange lines in Fig.\ref{fig:overview}), the neural discriminator takes in the generated trajectories and compares them with drivers' behavior data. Due to the blockage of gradients at the planning and control modules (blue dashed line in Fig.\ref{fig:overview}), we derive a novel learning objective to propagate credits of control actions back to the corresponding decisions (red line in Fig.\ref{fig:overview}). This learning framework is hence agnostic to the mechanisms of perception, planning, and control modules. We evaluate this framework on various simulated urban driving scenarios in CARLA, including (i) following, (ii) merging at the crossing, (iii) merging at the roundabout, and (iv) overtaking. The learning-based modular system demonstrates superior performance over rule-based modular systems and end-to-end control policies.

We summarize our contributions as:
\begin{itemize}
  \vspace{-8pt}
  \setlength{\itemsep}{0pt}
        \setlength{\parsep}{0pt}
        \setlength{\parskip}{0pt}
  \item The proposed framework combines a program-based system with a generative learning method, where the high-level decision policy is data-driven while the low-level planning and control modules remain configurable and physically constrained. These low-level constraints also improve sample efficiency in interactive learning.
  \item The proposed generative adversarial learning method is weakly supervised. It learns high-level driving decisions from low-level control data in an end-to-end manner, circumventing the ambiguous annotation of human driving decisions.
  \item This framework successfully distills behaviors in different scenarios into one decision module. Empirical results demonstrate that the learned decision module forms synergy with the programmed downstream modules, endowing the vehicle with smoother driving trajectories. It also exhibits substantial generalization capability in novel and complex scenarios.
        %\item \bz{Should we add this point: This framework successfully distillates the behaviors in different scenarios into one decision policy, which generalizes to unseen scenarios.}
\end{itemize}

\vspace{-5pt}
\section{Related Work}
\vspace{-5pt}
\label{sec:related}
\textbf{Classical autonomous driving system. } The design of the autonomous driving pipeline could be traced back to DARPA Urban Challenge 2007 \cite{buehler2009darpa}. The survey from \cite{paden2016survey} provides an overview of the hierarchical structure of perception~\cite{9196556,9123682}, decision~\cite{Sun_2020_corl}, planning, and control. Specifically, finite state machine (FSM) was used in the decision level, or equivalently upper planning level, in \cite{montemerlo2008junior}. Classical autonomous driving systems are organized in this way such that it is accessible to testing and diagnosis. Based on this modular pipeline, our work aims at developing a data-driven decision module to replace the FSM.

\textbf{Imitation learning-based autonomous driving system. } With the popularity of deep generative learning, methods have been proposed to learn end-to-end neural control policies from the driver's data. Previously, Ziebart \textit{et al.} \cite{ziebart2008maximum} and Ross \textit{et al.} \cite{ross2011reduction} proposed general methods in Inverse Reinforcement Learning and Interactive Learning from Demonstration, with an empirical study on a driving game. More recently, Kuefler \textit{et al.} \cite{kuefler2017imitating} and Behbahani \textit{et al.} \cite{behbahani2018learning} learn an end-to-end policy in a GAIL\cite{ho2016generative}-like manner. Codevilla \textit{et al.} \cite{codevilla2018end} and Liang \textit{et al.} \cite{liang2018cirl} share similar hierarchical perspective as us, but their control policies are completely neural. One thing worth mentioning is Codevilla \textit{et al.} \cite{codevilla2018end} proposed to learn a downstream policy through the imitation of human's high-level command, whose motivation is similar to ours. Rhinehart \textit{et al.} \cite{rhinehart2018deep} propose to learn a generative predictor for vehicles' coordinate sequence. Li \textit{et al.} \cite{li2018oil} imitate the mapping from predicted waypoints to control commands via neural network policy. Our work differentiates from all of them as we imitate the driver's low-level behaviors to learn high-level decisions, whose execution is conducted by a transparent and configurable program.

\textbf{Learning human intention. } Baker \textit{et al.} \cite{baker2009action} formalized humans' mental model of planning with Markov Decision Process (MDP) and proposed to learn human's mental states with Bayesian inverse planning. A similar probabilistic model of decisions is adopted by us but the learning is done with a novel generative adversarial method. Wang \textit{et al.} \cite{wang2013probabilistic} proposed a Hidden Markov Model (HMM) for human behaviors and materialized it with Gaussian Process (GP), where the intent is the hidden variable. Jain \textit{et al.} \cite{jain2015car} model the interaction between human intentional maneuvers and their behaviors in a driving scenario, both in-car and outside, as an Auto-Regressive Input-Output HMM and learn it with EM algorithm. However, they didn't explicitly distinguish humans from the environment. Besides, rather than inferring the human's mental state, we focus more on how a robot, \textit{i.e.} can make good decisions.
% execute these learned decisions.

%%%%%%%%%%%%%%%%%%%%%%%%%%%%%%%%%%%%%%%%%%%%%%%%%%%%%%%%%%%%%%%%%%%%%%%%%%%%%%%%

\vspace{-5pt}
\section{Modular Driving System}
\vspace{-5pt}
\subsection{System Overview}
\label{sec:overview}

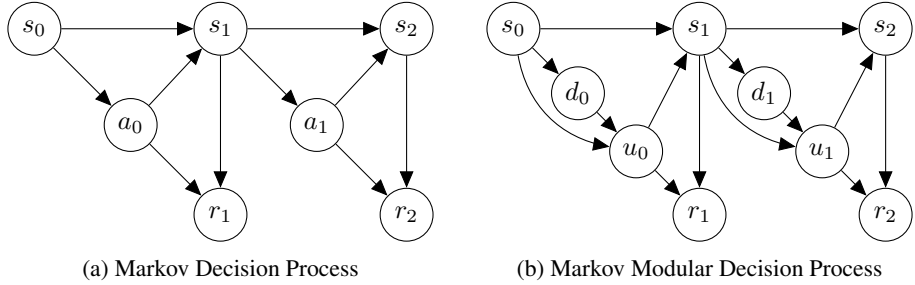
\begin{figure}
  \vspace{-0.1in}
  \centering
  \subfloat[][Markov Decision Process]{
    \begin{tikzpicture}
      \node[latent](s0){$s_0$};
      \node[latent, right=50pt of s0](s1){$s_1$};
      \node[latent, right=50pt of s1](s2){$s_2$};
      \node[latent, below right=22pt and 22pt of s0](a0){$a_0$};
      \node[latent, below right=22pt and 22pt of s1](a1){$a_1$};
      \node[latent, below=50pt of s1](r1){$r_1$};
      \node[latent, below=50pt of s2](r2){$r_2$};
      % \node[None, right=40pt of a1](cdots){$\cdots$};
      % \node[None, right=0pt of s2](cdots){$\cdots$};

      \draw[->, to path={-- (\tikztotarget)}]
      (s0) edge (s1);
      \draw[->, to path={-- (\tikztotarget)}]
      (s1) edge (s2);
      \draw[->, to path={-- (\tikztotarget)}]
      (s1) edge (r1);
      \draw[->, to path={-- (\tikztotarget)}]
      (s2) edge (r2);

      \draw[->, to path={-- (\tikztotarget)}]
      (s0) edge (a0);
      \draw[->, to path={-- (\tikztotarget)}]
      (a0) edge (s1);
      \draw[->, to path={-- (\tikztotarget)}]
      (a0) edge (r1);

      \draw[->, to path={-- (\tikztotarget)}]
      (s1) edge (a1);
      \draw[->, to path={-- (\tikztotarget)}]
      (a1) edge (s2);
      \draw[->, to path={-- (\tikztotarget)}]
      (a1) edge (r2);
    \end{tikzpicture}
    \label{subfig:mdp}
  }
  \hspace{1em}
  \subfloat[][Markov Modular Decision Process]{
    \begin{tikzpicture}
      \node[latent](s0){$s_0$};
      \node[latent, right=50pt of s0](s1){$s_1$};
      \node[latent, right=50pt of s1](s2){$s_2$};
      \node[latent, below right=10pt and 10pt of s0](d0){$d_0$};
      \node[latent, below right=10pt and 10pt of s1](d1){$d_1$};
      \node[latent, below right=32pt and 32pt of s0](u0){$u_0$};
      \node[latent, below right=32pt and 32pt of s1](u1){$u_1$};
      \node[latent, below=50pt of s1](r1){$r_1$};
      \node[latent, below=50pt of s2](r2){$r_2$};
      % \node[None, below right=-2pt and 45pt of d1](cdots){$\cdots$};
      % \node[None, right=0pt of s2](cdots){$\cdots$};

      \draw[->, to path={-- (\tikztotarget)}]
      (s0) edge (s1);
      \draw[->, to path={-- (\tikztotarget)}]
      (s1) edge (s2);
      \draw[->, to path={-- (\tikztotarget)}]
      (s1) edge (r1);
      \draw[->, to path={-- (\tikztotarget)}]
      (s2) edge (r2);

      \draw[->, to path={-- (\tikztotarget)}]
      (s0) edge (d0);
      \draw[->, to path={-- (\tikztotarget)}]
      (d0) edge (u0);
      \draw[->, to path={-- (\tikztotarget)}]
      (u0) edge (s1);
      \draw [->] (s0) to [out=-80,in=170] (u0);
      \draw[->, to path={-- (\tikztotarget)}]
      (u0) edge (r1);

      \draw[->, to path={-- (\tikztotarget)}]
      (s1) edge (d1);
      \draw[->, to path={-- (\tikztotarget)}]
      (d1) edge (u1);
      \draw[->, to path={-- (\tikztotarget)}]
      (u1) edge (s2);
      \draw [->] (s1) to [out=-80,in=170] (u1);
      \draw[->, to path={-- (\tikztotarget)}]
      (u1) edge (r2);
    \end{tikzpicture}
    \label{subfig:mmdp}
  }
  \caption{Comparison between \emph{Markov Decision Process} and \emph{Markov Modular Decision Process}}
  \vspace{-15pt}
  \label{fig:mmdp}
\end{figure}

%Before diving into the learning method, 
Fig.\ref{fig:overview} illustrates the proposed learning framework.
%the system structure of an intelligent driving system, which is also the modeling of the generator in our proposed learning framework, as shown in. 
We focus on the decision-making module and its connection to the downstream planning and control modules. The \textit{state estimator} (usually a perception module) constructs a local map by processing the sensory data and combining them with prior knowledge from the map and rules. The \textit{decision} module then receives \textit{observation} and decides a legal local driving task that steers the car towards the destination. To complete this task, the \textit{planning} module searches for an optimal trajectory under enforced physical constraints. The \textit{control} module then reactively corrects any errors in the execution of the planned motion.

In our proposed framework, the \textit{decision} module is a conditional probability distribution parameterized by neural networks. The downstream \textit{planning} and \textit{control} module are encapsulated. Because they are not the main contribution of this work, we adopt minimal settings to stay focused on our learning frameworks. These capsulated modules could be replaced by arbitrary alternatives.
%Since interfaces are of great importance in a modular system, in the following we will also draw readers' attention to our designate efforts on them. 

\subsection{Decision Making Module}
\label{sec:dmp}
We assume that a local map centered at the ego vehicle could be constructed from either the state estimator or an offline global HD map. This local map contains critical driving information such as the routing guidance to the destination, legal lanes, lane lines, other vehicle coordinates, and velocity in the ego vehicle's surroundings, as well as ego vehicle's speed and acceleration.

The interaction between the decision module and the environment through its downstream is modeled as Markov Decision Process (MDP). MDP is normally represented as a tuple $\mathcal{M}=(\mathcal{S}, \mathcal{D}, \mathcal{P}, r, \rho\textsubscript{0}, \gamma)$, with a state set $\mathcal{S}$, a decision set $\mathcal{D}$, a transitional probability distribution $\mathcal{P}:\mathcal{S} \times \mathcal{D} \times \mathcal{S} \rightarrow \mathbf{R}$,
% \textsubscript{+}
a bounded reward function $r:\mathcal{S} \times \mathcal{D} \rightarrow \mathbf{R}$, an initial state distribution $\rho\textsubscript{0}:\mathcal{S} \rightarrow
  \mathbf{R}$, a discount factor $\gamma \in [0, 1]$ for infinite horizon. Decision policy $\pi_{\theta}: \mathcal{S} \times \mathcal{D}\rightarrow \mathbf{R}$ takes current state $s_{t} \in \mathcal{S}$ from local map and generates a high-level decision $d_{t}$. Note that the notation of $\mathcal{D}$ and $d$ are unconventional, we explicitly denote \textit{decision} with $d_{t}$ to differentiate it from \textit{control action} $u_{t}$. This decision directs the execution of planning and control module, and makes the autonomous driving system proceed in the environment to acquire next state $s_{t+1}$. Therefore there is a modification of the decision process, termed as Markov Modular Decision Process shown in Fig.\ref{fig:mmdp}. The optimization objective of this policy is to maximize the expected discounted return $\mathbf{E}_{\tau}[\sum_{t=0}^{T}\gamma^{t}\mathcal{R}(s_{t}, u_{t})]$, where $\tau = (s_0, d_0, u_0, ...)$ denotes the whole trajectory, $s_{0} \sim \rho_{0}(s_{0})$, $d_{t}\sim\pi_{\theta}(d_{t}|s_{t})$, $u_{t}\sim P(u_{t}|s_{t}, d_t)$ and $s_{t+1}\sim\mathcal{P}(s_{t+1}|s_{t},u_{t})$.

Below we introduce the \textit{observation} space and \textit{decision} space, which are the interfaces in our modular system. Note that we assume full access to necessary information on the local map.

\begin{wrapfigure}{r}{0.5\linewidth}
  \centering
  \includegraphics[width=\linewidth]{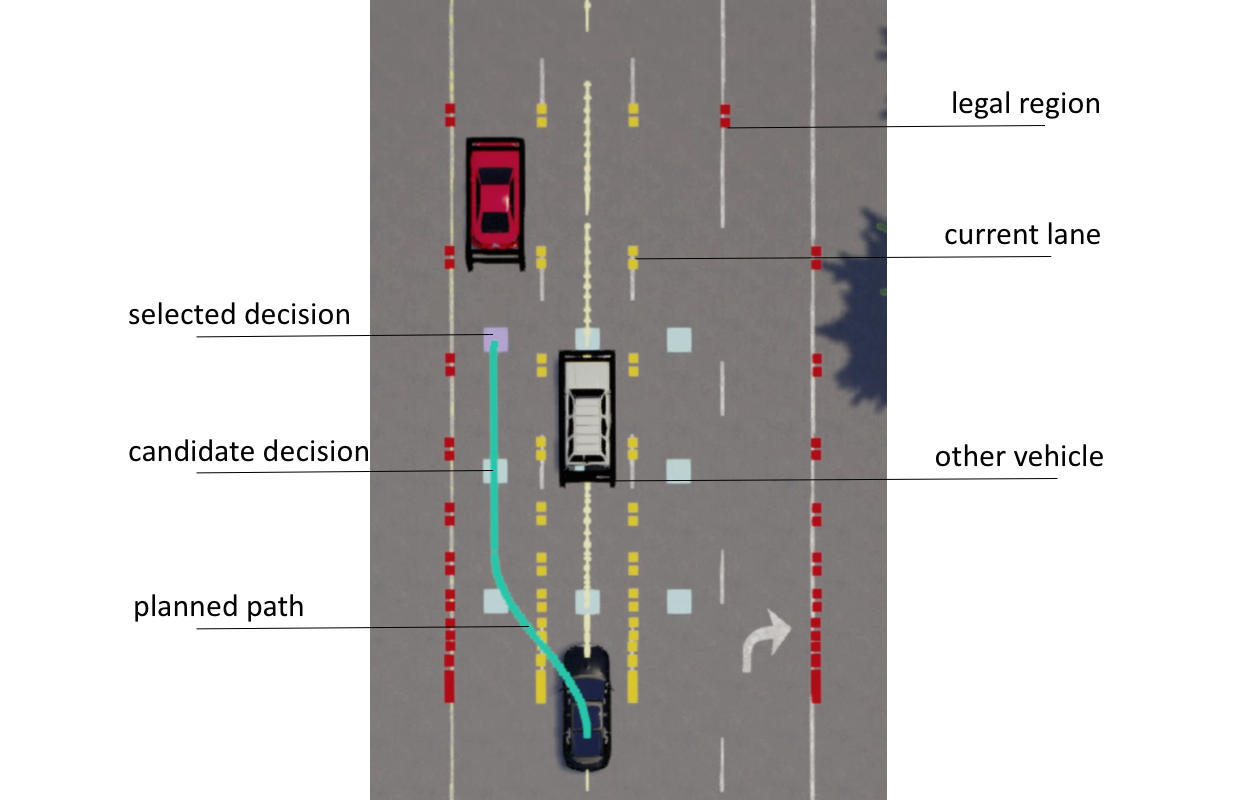}
  \caption{\textbf{Illustration of input and output of the decision module when the driving system conducts \textit{overtake}}. The trained policy decides to pick the left lane, in terms of the target point and the target speed. The planner generates a trajectory, which is tracked by the controller. Observations are shown on the right. Yellow lines are the ego vehicle's current lane lines and the red lines are the edge of \textit{legal} regions. }
  \label{fig:decision}
  \vspace{-10pt}
\end{wrapfigure}

To make the decision-making policy more generalizable to different driving scenarios, the output of the state estimator, also as the input to the neural decision module, is the local map and traffic state. As shown in Fig.\ref{fig:decision}, a part of the observation is the 3D coordinates of sample points of lane lines. We select two sets of lanes for the decision policy. One for the ego vehicle's current lane and the other for the edge of \textit{legal} regions. Note that here \textit{legal} means driving in that region abides by both the traffic rules and a global routing, which would have a crucial effect at the crossing. To reduce the dimension of input, these lane points are sampled with exponentially increasing slots towards the further end. In some sense, it mimics the effect of Lidar, with the observation more accurate in the near end. The observation also includes the coordinates and velocities of the $6$ nearest vehicles in traffic within $70m$ range of the ego vehicle.

% We introduce strong priors over the decision space for tractable inversion of this generation process 
%MDP models 
% from the collected data. 
We define the \textit{decision} as three independent categorical variables, one for lateral, one for longitudinal and one for velocity. Combining with the local map, each of them is assigned a specific semantic meaning. The lateral decision variable has three classes as \textit{changing to the left lane}, \textit{keeping current lane} and \textit{changing to the right lane}. Note that at a crossing, global routing information has been implanted into the local map as introduced in the last paragraph. The lateral decision set is complete since these three are the only possible lateral decisions for a vehicle. The longitudinal decision variable has four different classes, with each indicating \textit{the distance along with waypoints in time interval T}. The exact coordinate of the endpoint could be extracted from the local map. Combining with the predicted target speed at the endpoint from the velocity decision variable, which discretizes the allowed speed range equally into four baskets, this decision module provides a quantitative configuration of goal states for the downstream trajectory planner.

\subsection{Planning and Control Modules}
\label{sec:pnc}
We introduce a minimal design of the planning module and the control module for completeness, which is not the main contribution of this work. In the classical autonomous driving system, the planning module calculates the trajectories under constraints such as collision avoidance. Its optimization objective is to reach the goal state specified by the decision module with minimal cost. In our minimal setting, the planning module processes path and velocity separately. Paths are planned with cubic Bezier curves, while velocity is planned with Quadratic Programming (QP) to minimize the acceleration, jerk, and discontinuity in curvature. The control module drives the vehicle to trace the planned trajectories. And we implement it minimally with PID. Details are provided in Appx. A.

It is flexible to further extend each module of the proposed framework, as long as the alternatives are deterministic or stationary. Note that while the interface between decision and planning is the location of a goal point and the speed ego vehicle is expected to maintain when reaching it, the planning module could be replaced by a more sophisticated search-based or optimization-based trajectory planner, to fulfill more practical requirements in execution time and constraint enforcement. Similarly, the model-free controller could be replaced with alternatives like Model Predictive Control (MPC). Other practical constraints and concerns over the controller such as robustness and stability could also be taken into consideration if necessary.

\section{Imitation Learning}
\label{sec:learning}
\vspace{-5pt}
\subsection{From IRL to GAIL}
Under the MDP modeling, we choose imitation learning over reinforcement learning. In principle, imitation learning could save practitioners from \textit{ad hoc} reward engineering and focus their effort on reusable infrastructure development. Among all the imitation learning methods, behavior cloning (BC) \cite{pomerleau1991efficient} is the most straightforward one. The policy is trained as a regressor in a purely supervised manner, ignoring the temporal effect of each action in an MDP trajectory. Thus it suffers from covariate shift when presented with states which are not covered by the training data.
%If the transition probability, aka the model of the environment, is analytically known, the temporally accumulated shift could be anticipated and thus mitigated \cite{sun2017deeply}. But it is normally not the case. 
By contrast, imitation learning such as Inverse Reinforcement Learning (IRL) explicitly models the interaction between policy and environment and approximates for exact value iteration \cite{ziebart2008maximum} or Monte Carlo approximation \cite{ho2016generative}. Thus they can generalize better in states not covered in the demonstration.

Fig.2 illustrates the difference in the graphical model of a trajectory from an MDP when the modular system is adopted for the agent. In this modified probabilistic model, a trajectory is a sequence $\tau = (s_0, d_0, u_0, s_1, d_1, u_1...)$. However, what is observable from demonstration is $\hat{\tau} = (s_0, u_0, s_1, u_1...)$. According to the formulation of MaxEnt IRL \cite{ziebart2008maximum}, when maximum entropy principle is applied for the tie-breaking between cost functions that can generate identical optimal trajectories, the probability of a trajectory is
\begin{equation}
  p(\hat{\tau}|R) = \frac{1}{Z}\exp(R(\hat{\tau})),
\end{equation}
where $Z$ is the partition function, $R$ is the reward function, which is normally Markovian $R(\hat{\tau})=\sum r(s_i, u_i)$.
Given a set of demonstration $\mathcal{D}_{E}$, we can infer the reward function by Maximum Likelihood (MLE):
\begin{equation}
  \mathcal{L}_{IRL}=\mathbf{E}_{{\mathcal{D}}_{E}}[\log p(\hat{\tau}|R)] = \mathbf{E}_{{\mathcal{D}}_{E}}[\sum_{\hat{\tau}} r(s_i, u_i) - \log Z].
\end{equation}
In the standard IRL, with the learned reward function, we can solve for the policy with any reinforcement learning methods. However, as the partition function is always intractable if the state space is continuous, methods have been proposed to approximate it \cite{ziebart2008maximum, finn2016guided}. Basically, there is a background distribution $q(\hat{\tau})$ for the estimation of $Z$. In \cite{finn2016connection}, it is further illustrated how a delicately designed importance sampling scheme can connect this IRL loss to a specific type of discriminator over trajectories. Ho \textit{et al.} \cite{ho2016generative} propose Generative Adversarial Imitation Learning (GAIL) to further approximate this discriminator with a step-wise one, remedy the trajectory-wise information with advantage accumulation and learn a policy with off-the-shelf reinforcement learning methods. Intuitively, it learns a policy whose actions at given states are indistinguishable from demonstration data. More formally, with expert demonstration ${\mathcal{D}}_{E}$, a neural discriminator $D_{\phi}:\mathcal{S} \times \mathcal{U} \rightarrow (0, 1)$ is introduced and the reward function is defined as $r(s, u)=-\log(D_{\phi}(s, u))$. The training objective is:
\begin{equation}
  \begin{split}
    \mathcal{L}_{GAIL}=\min_{\theta} \max_{\phi}\{\mathbf{E}_{p_{\theta}(u|s)}[\log(D_{\phi}(s,u))]
    +\mathbf{E}_{{\mathcal{D}}_{E}}[\log(1-D_{\phi}(s^{E},u^{E}))]\}.
    \label{eq:gail}
  \end{split}
\end{equation}
$D_{\phi}$ is optimized with cross entropy loss, while $\pi_{\theta}$ is optimized with policy gradient \cite{schulman2015trust}.
\vspace{-5pt}
\subsection{Handling Non-differentiable Downstream Modules}
\vspace{-5pt}
Different from GAIL, in our framework the generation process is modularized, following the structure of classical autonomous driving systems. Because the downstream planning and controls modules are not necessarily differentiable, a separation occurs between decisions from neural policy and the control data from drivers, as illustrated by the blue dotted line in Fig 1: policy $\pi_{\theta}$ generates \textit{decisions} $d_{t}$ while discriminator only distinguishes \textit{actions} $u_{t}$. And as shown in Fig 2, $p_\theta (u_t|s_t) = \sum_{d_t}p(u_t|d_t)\pi_{\theta}(d_t|s_t)$. Since planning and control modules are both deterministic, at every state, once the decision is chosen, the transformation from decision to control is a deterministic and correspondence mapping. Hence, one important insight of our work is that the transformation at each state $s$
\begin{equation}
  \begin{split}
    \label{eq:push-forward}
    g_s:\{d_t\} \xrightarrow[\text{1-to-1}]{\text{onto}}  \{u_t\},
  \end{split}
\end{equation}
which is the abstraction of the planning and control module, is a deterministic and correspondence mapping.
Therefore, it is a candidate for \textit{reparametrization} \cite{kingma2013auto} or \textit{push-forward} \cite{rhinehart2018r2p2}. Reparametrize $u$ in the first expectation in Eq.\ref{eq:gail} with $d$:
\begin{equation}
  \begin{split}
    \label{eq:reparam}
    &\mathbf{E}_{p_{\theta}(u|s)}[\log(D_{\phi}(s,u))]=\sum p_{\theta}(u|s)\log(D_{\phi}(s,u))\\
    =& \sum {\pi_{\theta}(d|s)}\log(D_{\phi}(s,g_s(d))) = \mathbf{E}_{\pi_{\theta}(d|s)}[\log(D_{\phi}(s,g_s(d)))].
  \end{split}
\end{equation}
Intuitively, we only use this black-box function $g_s$ for Monte Carlo sampling, thus there is no need to know its analytical form. In the training stage, we do not actually need to calculate $u_t$ with the push-forward function as in Eq.\ref{eq:push-forward}. Instead, we calculate the true $g_s(d_t)$ according to Sec. 3 during sampling, and save both $d_t$ and $u_t$ for training. During training, $d_t$ is fed to $\pi_\theta$ and the corresponding $u_t$ is fed to $D_\phi$. Since $\mathcal{D}$ is discrete, it is differentiated with policy gradient. We then have this modified learning objective:
\begin{equation}
  \begin{split}
    \mathcal{L}=\min_{\theta} \max_{\phi}\{\mathbf{E}_{\pi_{\theta}(d|s)}[\log(D_{\phi}(s,u))]
    +\mathbf{E}_{{\mathcal{D}}_{E}}[\log(1-D_{\phi}(s^{E},u^{E}))]\}.
  \end{split}
\end{equation}
The whole algorithm is described in Algorithm \ref{alg:main}.

\begin{minipage}{.54\linewidth}
  \vspace{0pt}
  \begin{algorithm}[H]
    \caption{Learning executable decisions from human behavioral data}
    \label{alg:main}
    \begin{algorithmic}
      \REQUIRE Expert demonstration ${\mathcal{D}}_{E}$, batch size $B$
      \STATE Initialize policy $\pi_{\theta}$, discriminator $D_{\phi}$, (optionally) value function $V_{w}$
      \WHILE{policy iteration not converged}
      \STATE Initialize data buffer $\mathcal{B}$ with length $B$
      \WHILE{$\neg \mathcal{B}.full()$}
      \STATE Sample expert trajectory: $\tau_{E} \sim {\mathcal{D}}_{E}$
      \STATE Initialize $sim$ with $s_{0}^{E}$
      \STATE $traj$ = GENERATE-TRAJ
      \STATE Append $traj$ to $\mathcal{B}$
      \ENDWHILE
      \STATE Compute reward for samples in $\mathcal{B}$ with $D_{\phi}$
      \STATE Update $D_{\phi}$ with data in $\mathcal{B}$ and ${\mathcal{D}}_{E}$
      \STATE Policy iteration on $\pi_{\theta}$ with tuples in $\mathcal{B}$, (optionally with value iteration on $V_{w}$)
      \ENDWHILE
    \end{algorithmic}
  \end{algorithm}
\end{minipage}
\hspace{5pt}
\begin{minipage}{.42\linewidth}
  \vspace{0pt}
  \begin{algorithm}[H]
    \caption{GENERATE-TRAJ}
    \label{alg:gen}
    \begin{algorithmic}
      \REQUIRE Simulator (or ROS offline replayer) with local map $sim$, planning module $planner$, control module $controller$, maximum trajectory horizon $\mathcal{H}$
      \STATE Initialize placeholder $traj$, $t=0$
      \WHILE{$t<\mathcal{H}$ or $\neg sim.done$ }
      \STATE Get $s_{t}$ from $sim$
      \STATE Sample $d_{t}$ from $\pi_{\theta}(o_{t})$
      \STATE Get $pln_{t}$ from $planner(sim, d_{t})$
      \STATE Get $u_{t}$ from $controller(sim, pln_{t})$
      \STATE Append $(s_{t}, d_{t}, u_{t})$ to $traj$
      \STATE Send $u_{t}$ to $sim$
      \STATE $t++$
      \ENDWHILE
      \STATE Yield $traj$
    \end{algorithmic}
  \end{algorithm}
\end{minipage}
\section{Experiments}
\begin{figure*}[tp]
  \centering
  \vspace{-15pt}
  \begin{subfigure}{.23\textwidth}
    \centering
    % include second image
    \includegraphics[width=0.95\linewidth]{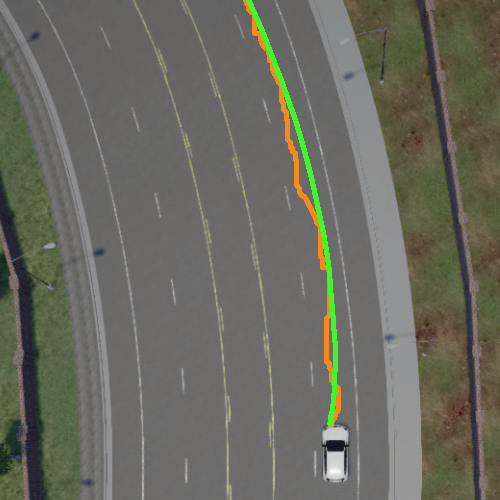}
    %   \caption{Two-lane\_Car\_Following}
    \label{subfig:smooth-1}
  \end{subfigure}
  \begin{subfigure}{.23\textwidth}
    \centering
    % include first image
    \includegraphics[width=0.95\linewidth]{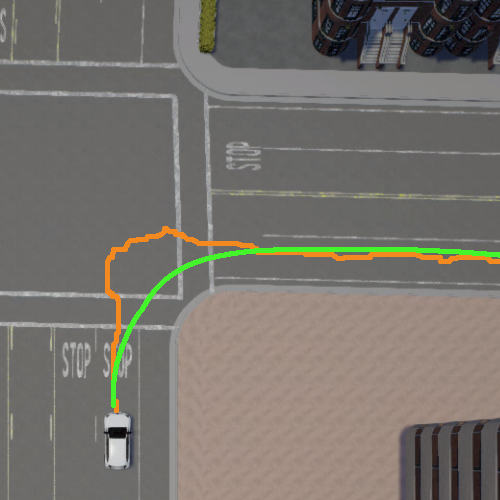}
    %   \caption{Crossroad\_Merge}
    \label{subfig:smooth-2}
  \end{subfigure}
  \begin{subfigure}{.23\textwidth}
    \centering
    % include first image
    \includegraphics[width=0.95\linewidth]{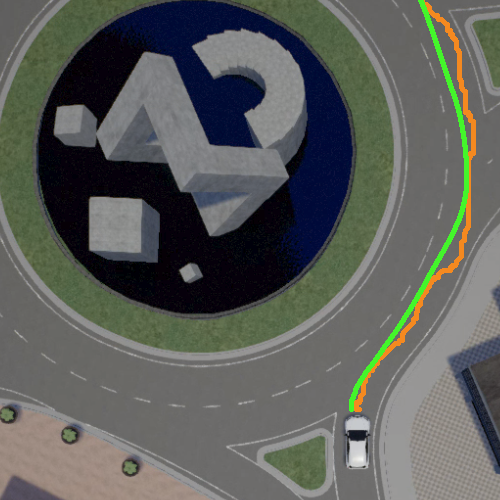}
    %   \caption{Roundabout\_Merge}
    \label{subfig:smooth-3}
  \end{subfigure}
  \begin{subfigure}{.23\textwidth}
    \centering
    % include first image
    \includegraphics[width=0.95\linewidth]{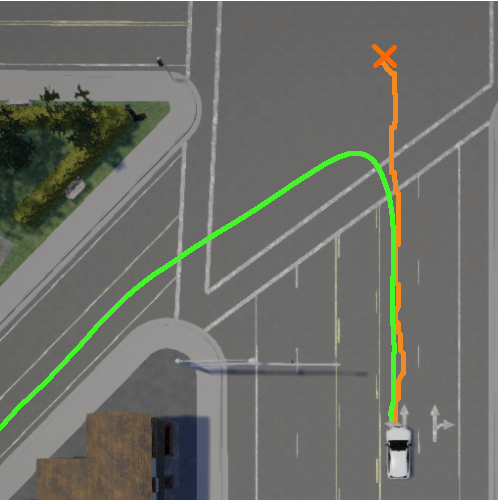}
    %   \caption{Crossroad\_Turn\_Left}
    \label{subfig:smooth-4}
  \end{subfigure}
  \caption{Trajectories from our proposed framework (green lines) are smoother than the ones from the end-to-end neural policy (orange lines), because of the explicit enforcement of geometrical constraints in the downstream planning module. ($\times$ indicates the car violates routing rules.)}
  \vspace{-10pt}
  \label{fig:smooth}
\end{figure*}

%We believe results from these rudimentary scenarios could be generalized to others
%Configuration details of these scenarios are listed in Appendix. 
% \begin{figure*}
% \centering

% \caption{Crossroad\_Turn\_Left Scenario}
% \label{fig:crossroad-turn-left}
% \end{figure*}

% \begin{figure*}
% \centering

% \caption{Roundabout\_Merge Scenario}
% \label{fig:roundabout-merge}
% \end{figure*}
\vspace{-5pt}
\subsection{Rudimentary Driving in Empty Town}
\vspace{-5pt}
%\emph{Empty Town} is a special traffic scenario where there is no traffic. It covers urban road structures like a straight road, curve road, crossing, and roundabout. 
Experiments in \emph{Empty Town} show the difference between a modular driving system and an end-to-end neural control policy. The training time on an $8$-GPU computation server for the end-to-end control policy is $31$ hours versus $15$ hours for our modular driving system. Different training time in the same platform implies different numbers of interactions these two methods need to converge, showing the advantage of using the modular pipeline. %in exploration from constraint enforcement. 

As shown in Fig.\ref{fig:smooth}, trajectories from our proposed framework are smoother than the ones from the end-to-end policy. This demonstrates the effect of explicit enforcement of geometrical constraints in the downstream planning module. Interestingly, there are some road structures where the learning-based modular system can pass while the end-to-end control policy cannot. For example, Fig.\ref{fig:smooth}(d) shows a narrow sharp turn where end-to-end control policy drives across the legal lane line. We believe this is a difficult temporally-consistent exploration problem. An agent needs some successful trials of consecutive right-turning action samples to learn, which is challenging for generative interactive learning. The modular system, in contrast, only explores behaviors that are temporally plausible like human drivers. With the planning module that searches for a geometrically constrained path, the agent can effortlessly make this sharp turn.

% For the experiments in \emph{Empty Town}, we filtered out $75$ cases among $300$ testing points given by CARLA. 
Table \ref{tab:quant-comparison} shows the statistics of behaviors from the learning-based modular system, end-to-end control policy, and rule-based method. The comparison shows that our method can drive more safely ($0\%$ collision rate), much faster (less time to finish), and with higher comfort (less acceleration). This exhibits the advantage of having both the learned decision policy and the classic planning and control module.
% \jh{Note that the end-to-end control policy only fails the narrow sharp turn case (Fig. \ref{fig:smooth}(d)) in all 75 testing cases. Hence the collision rate for end-to-end policy in \emph{Empty Town} is 1.33\%.} 
Besides, the decision module trained in Town 3 works similarly well in Town 2, showing its generalization ability.
\vspace{-5pt}
\subsection{Socially Interactive Driving in Traffic Scenarios}
\vspace{-5pt}
We further test our model's performance in each traffic scenario. The statistics of Traffic Scenarios in Table \ref{tab:quant-comparison} are the average of $100$ evaluations.
Traffic scenarios include basic scenarios such as \emph{Car Following} and \emph{Crossroad Merge} where vehicle's interactions with zombies are relatively simple and monotonic; and complex ones such as \emph{Crossroad Turn Left}, \emph{Roundabout Merge} and \emph{Overtake}. In complex traffic scenarios, the dynamics around the ego vehicle is of higher variation, due to either more traffic users or more complex relations between them.

As shown in Table \ref{tab:quant-comparison}, in basic traffic scenarios,
% all three methods obtain agents that can drive safely (0 collision rate). Among them,
rule-based and learning-based modular systems are somehow on par in terms of time taken to finish, while rule-based agent seems to drive a little bit more comfortably. End-to-end agent drives more rudely in most of the scenarios, expect in \textit{Crossroad Merge}, where it takes much longer time to finish. Since the modular pipeline enforces smoother planning and control, the learning-based modular system offers more driving comfort than the end-to-end system. In basic traffic scenarios, none of them drives as well as experts.

In complex traffic scenarios,
%  rule-based agent configured in two sub-scenarios fails to drive safely in another one.
rule-based agent configured in some scenarios fails to drive safely in others while learning-based modular agent is safer and smarter. This is because rule-based decision module is based on FSM, which is composed of a huge number of rules. In order to drive safely in all environments, the rules of safety have priority over other factors such as driving comfort. In complex environments, the rules in FSM would be more conservative to handle corner cases and uncertainty(\emph{e.g.}, more hard brakes or accelerations). Unlike the rule-based system, the learning-based system can minimize the acceleration and jerk by imitating human experiences. Furthermore, since the rule-based system is program-based, its configurations might vary in different scenarios, which makes it difficult to generalize across scenarios.
%  , just as in basic traffic scenarios. 
When compared with end-to-end control policy, a similar conclusion could be drawn as in the previous subsection that learning-based modular agent offers more comfort because the modular pipeline enforces smoother planning and control. Interestingly, in complex social scenarios, the learning-based modular system achieves higher comfort (lower acceleration) than experts. This may be attributed to human errors, and a learning-based modular agent somehow fixes it. Experiments of generalization capability can be found in Appendix.

% \usepackage{multirow}
\begin{table*}[tp]
  \centering
  \vspace{-15pt}
  \caption{Performance Comparision (LBM: Learning-based Module, E2E: End-to-end neural policy, RB: Rule-based method, ED: Expert Data)}
  \label{tab:quant-comparison}
  % \begin{tabular}{>{\centering\arraybackslash}m{0.10\linewidth}>{\centering\arraybackslash}m{0.04\linewidth}>{\centering\arraybackslash}m{0.04\linewidth}>{\centering\arraybackslash}m{0.04\linewidth}>{\centering\arraybackslash}m{0.04\linewidth}>{\centering\arraybackslash}m{0.04\linewidth}>{\centering\arraybackslash}m{0.04\linewidth}>{\centering\arraybackslash}m{0.04\linewidth}>{\centering\arraybackslash}m{0.04\linewidth}>{\centering\arraybackslash}m{0.04\linewidth}>{\centering\arraybackslash}m{0.04\linewidth}>{\centering\arraybackslash}m{0.04\linewidth}>{\centering\arraybackslash}m{0.04\linewidth}>{\centering\arraybackslash}m{0.04\linewidth}}
  \begin{tabular}{cccccc}
    \toprule[1pt]
    % \multirow{2}{*}[-0.15in]{\textbf{Scenarios}}   & \multicolumn{4}{c}{\textbf{Learning-based Module}}                                     & \multicolumn{4}{c}{\textbf{End-to-End}}                       & \multicolumn{4}{c}{\textbf{Rule-based}} & \textbf{Expert Data}                   \\ 
    % \cmidrule(lr){2-5} \cmidrule(lr){6-9} \cmidrule(lr){10-13} \cmidrule(lr){14-14} 
    %                              &\textbf{CL} & \textbf{Time Taken ($s$)}         & \textbf{Accel. ($m/s^2$)} & \textbf{Jerk ($m/s^3$)}  & \textbf{CL)} & \textbf{Time Taken ($s$)} & \textbf{Accel. ($m/s^2$)} & \textbf{Jerk ($m/s^3$)} & \textbf{CL} & \textbf{Time Taken ($s$)} & \textbf{Accel. ($m/s^2$)} & \textbf{Jerk ($m/s^3$)} & \textbf{Accel. ($m/s^2$)}\\ 
    % \multicolumn{2}{c}{\textbf{Scenarios}}   
    \textbf{Scenario}                                                                              & \textbf{Time Taken ($s$)}                                                                  & \textbf{Acceleration ($m/s^2$)}                                                           & \textbf{Jerk ($m/s^3$)}                                                                            & \textbf{Collision Rate ($\%$)}                                                       \\
    \toprule[1pt]
    \rotatebox{90}{\emph{Empty Town}}                                                              &
    \includegraphics[width=0.14\linewidth]{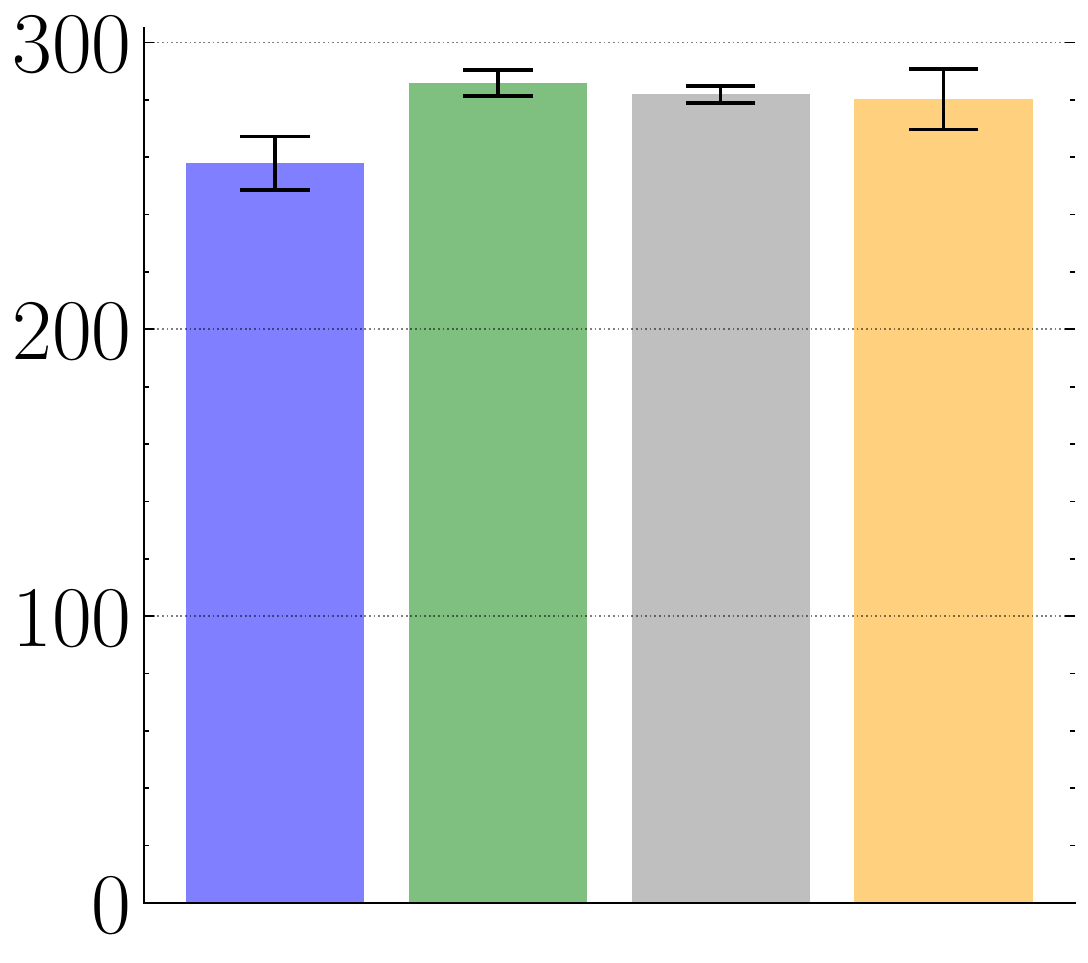}              & \includegraphics[width=0.14\linewidth]{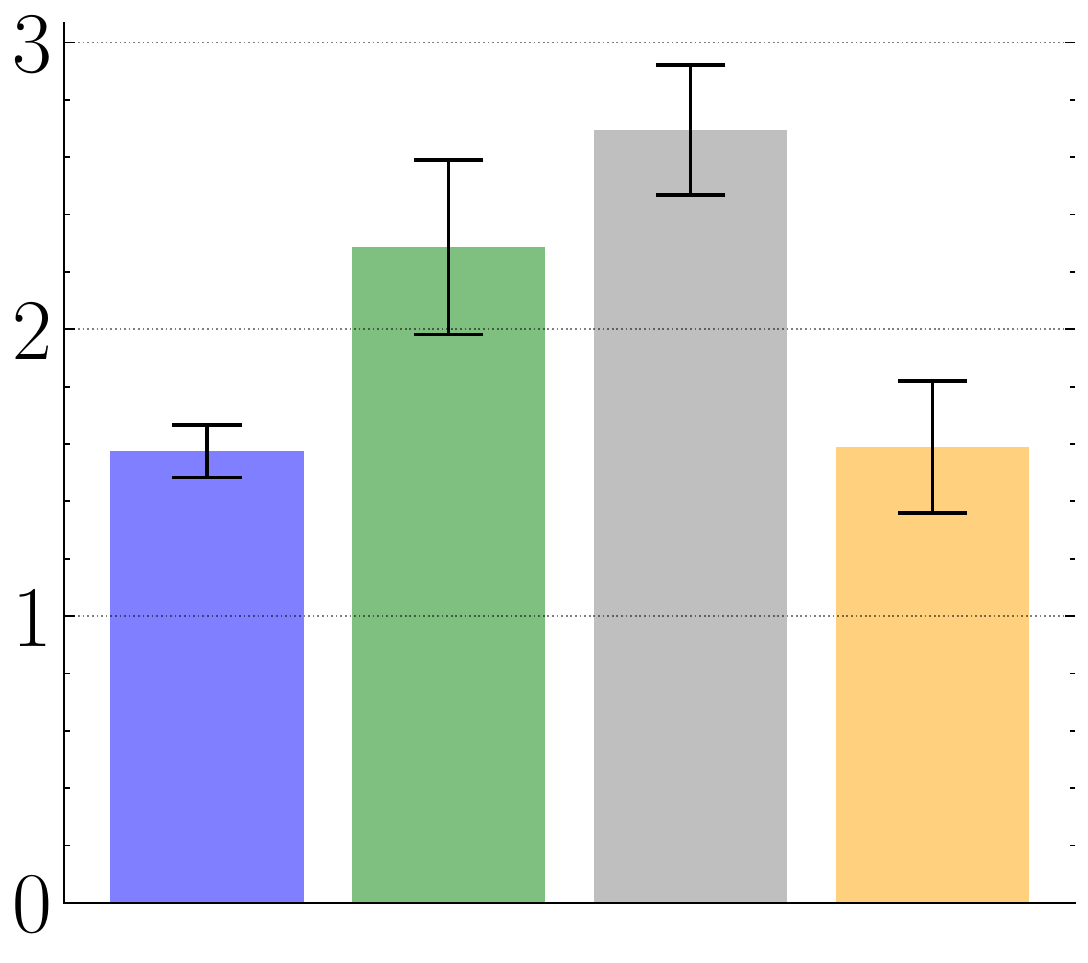}              & \includegraphics[width=0.14\linewidth]{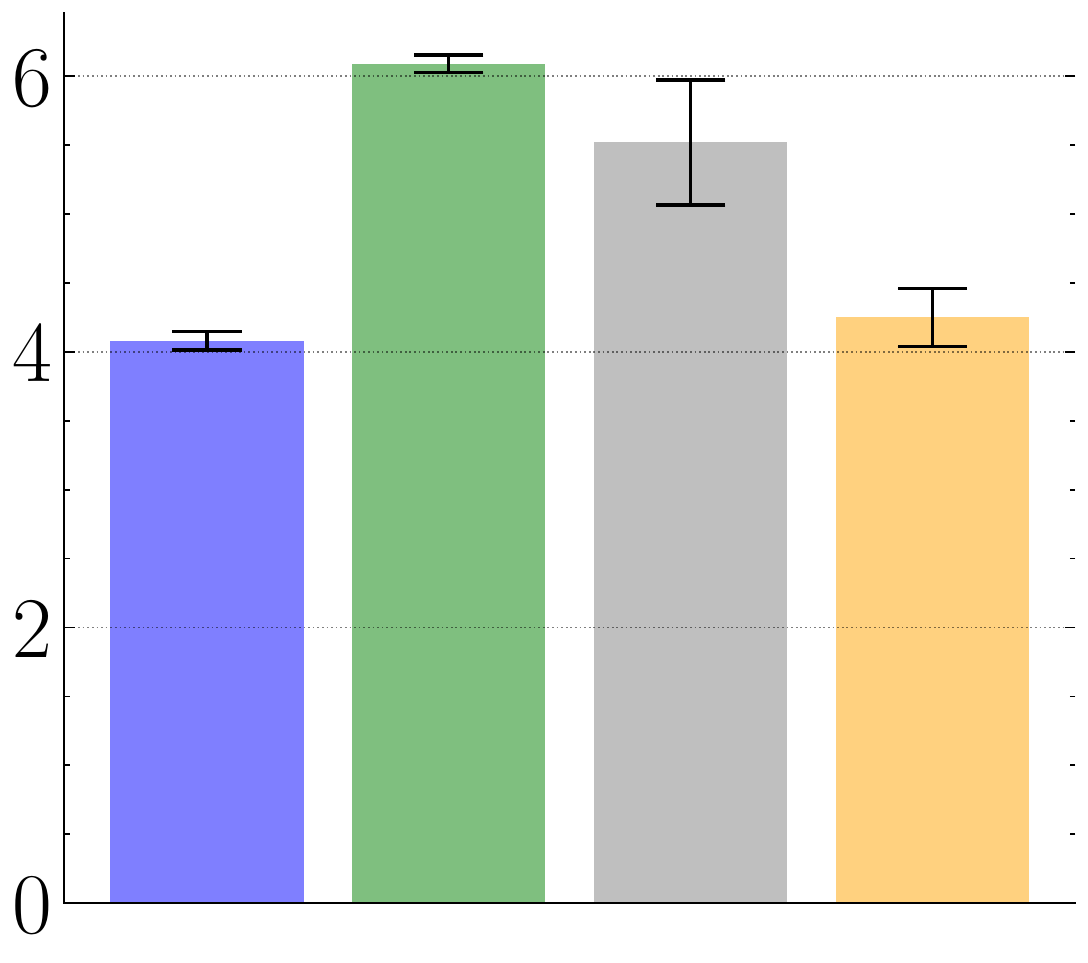}              & \includegraphics[width=0.14\linewidth]{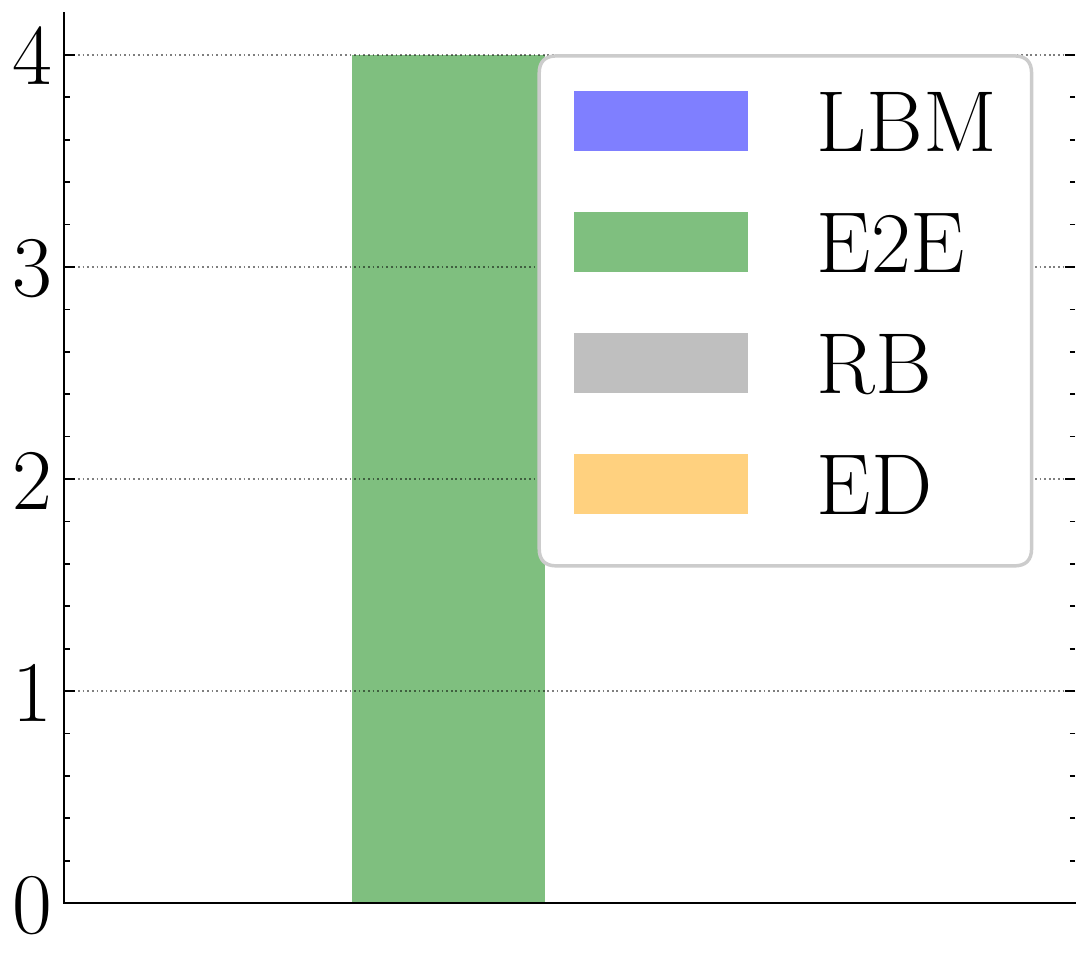}
    \\ \hline
    \rotatebox{90}{\parbox{0.14\linewidth}{\emph{Two                                                                                                                                                                                                                                                                                                                                                                                                                                    \\ Lanes  Car \\ Following}}}   &
    \includegraphics[width=0.14\linewidth]{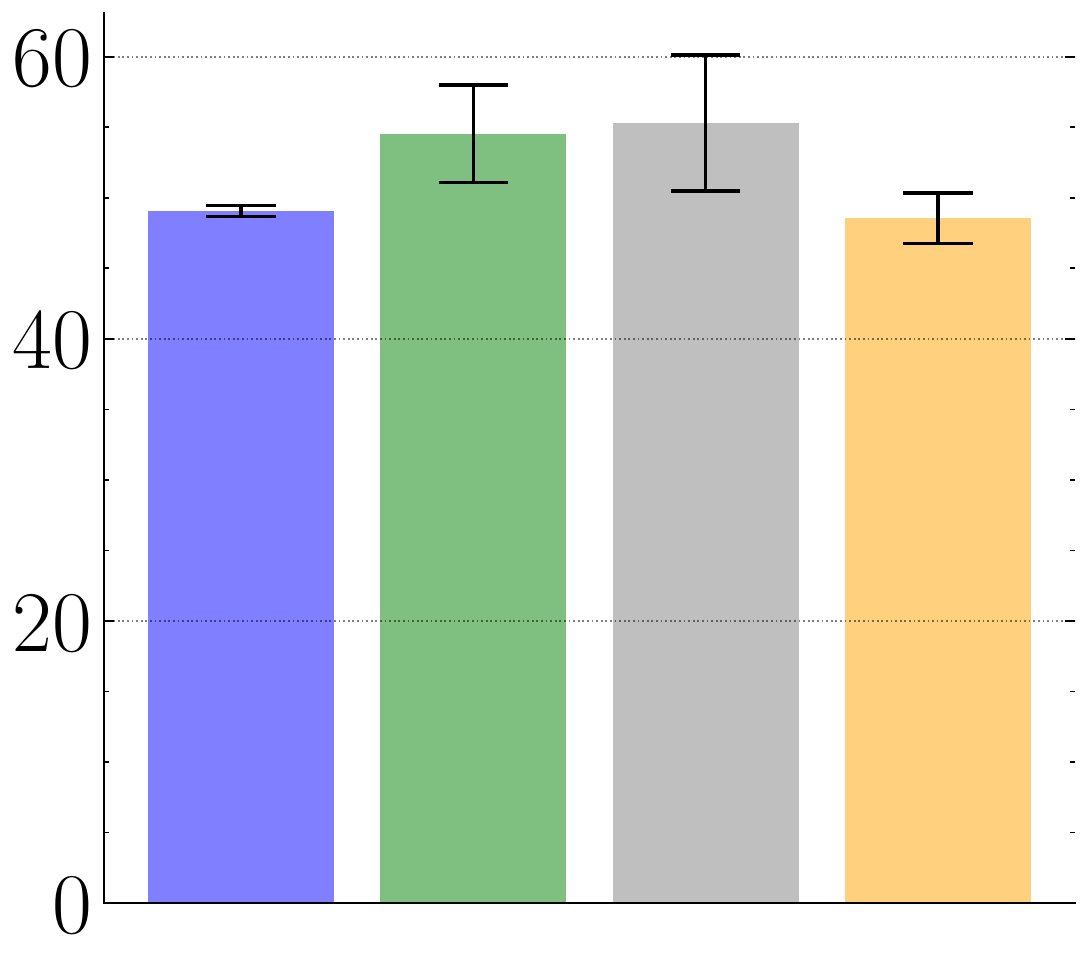}   & \includegraphics[width=0.14\linewidth]{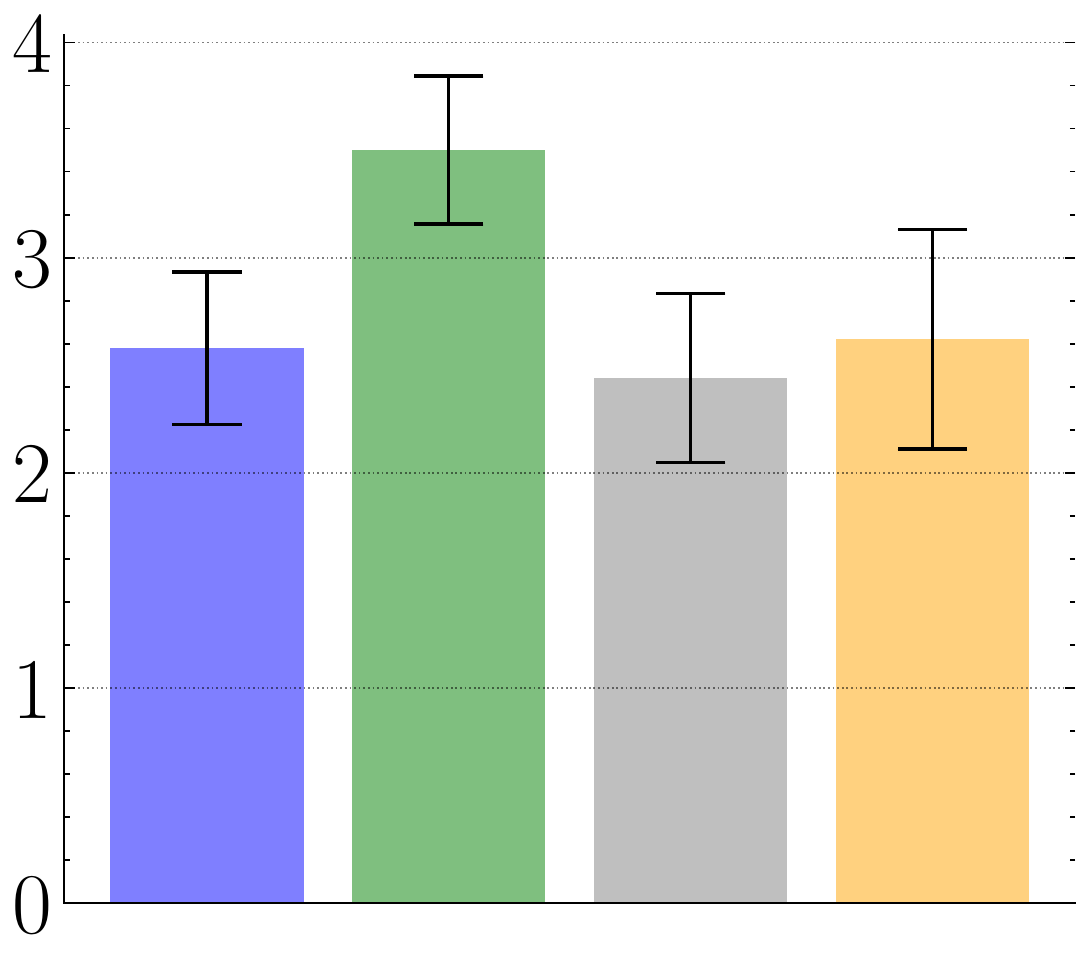}   & \includegraphics[width=0.14\linewidth]{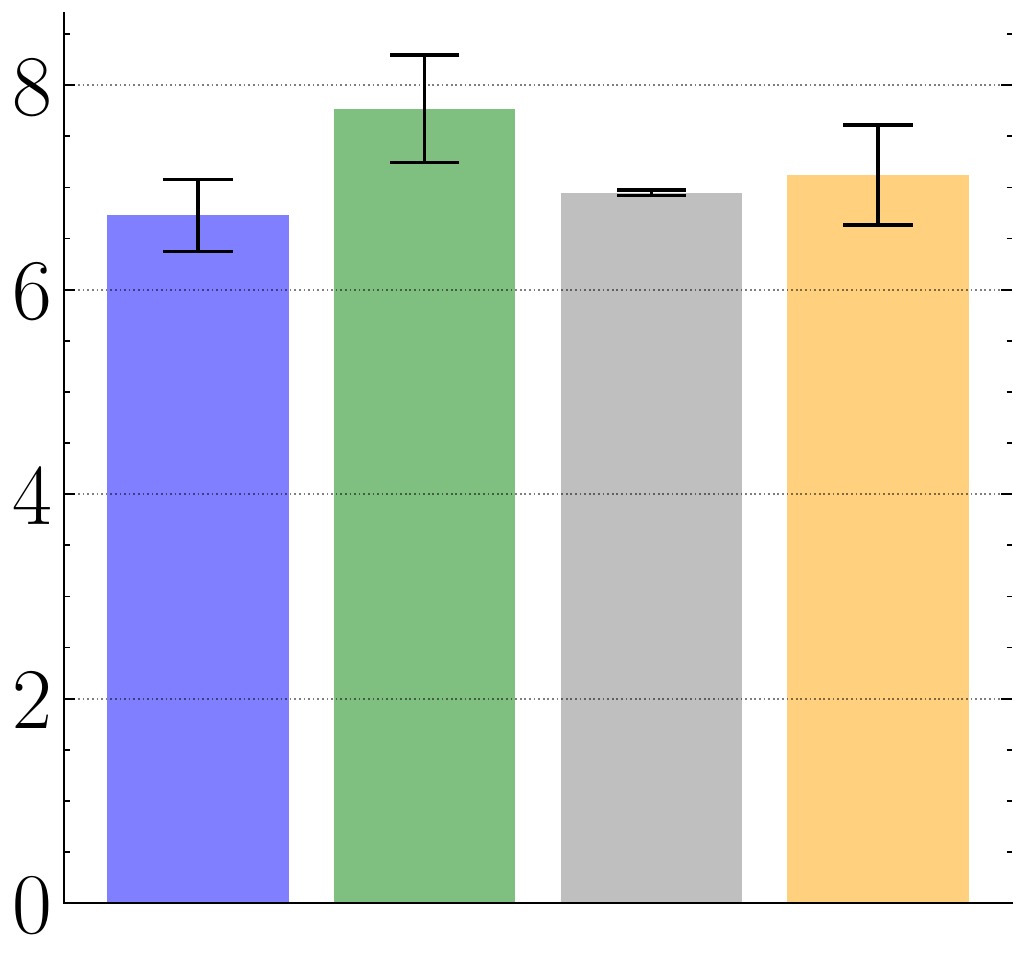}   & \includegraphics[width=0.14\linewidth]{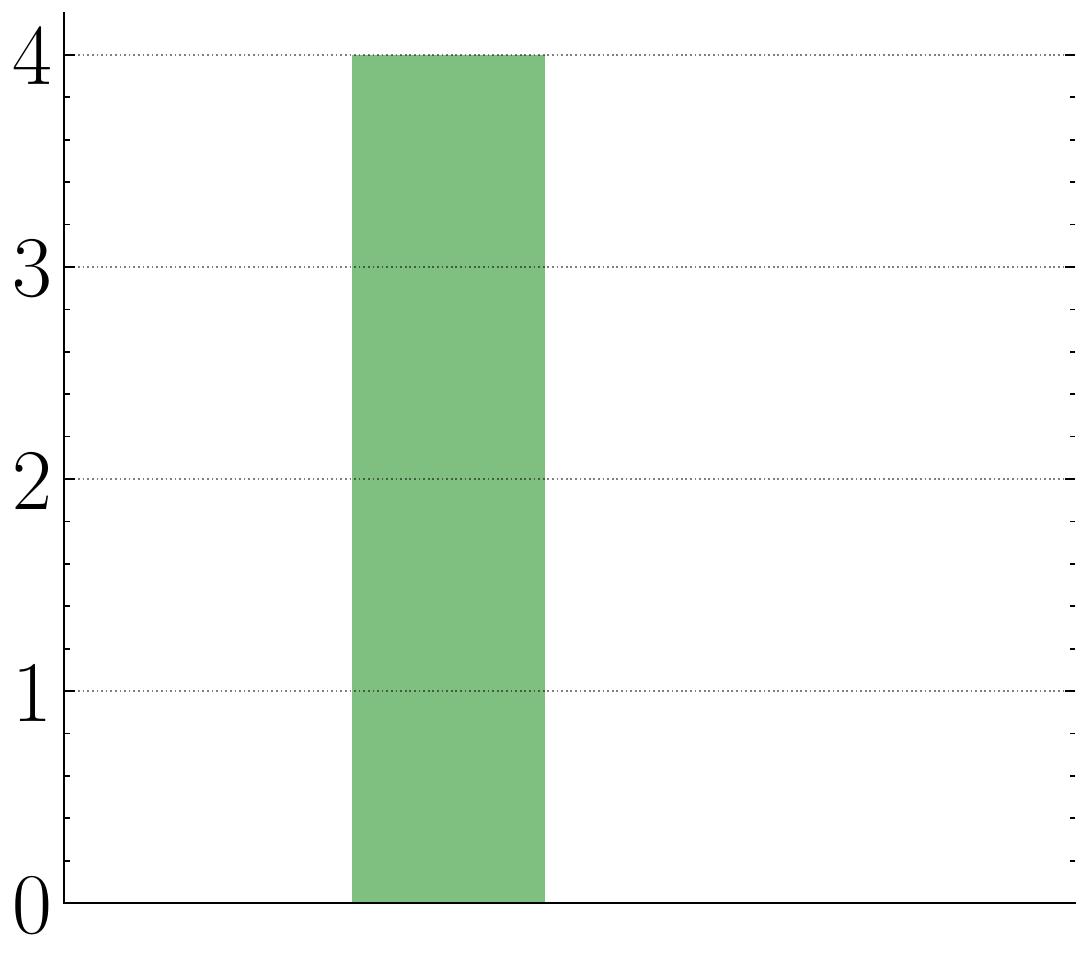}                                                                                          \\ \hline
    \rotatebox{90}{\parbox{0.14\linewidth}{\emph{Single                                                                                                                                                                                                                                                                                                                                                                                                                                 \\ Lane Car \\ Following}}} &
    \includegraphics[width=0.14\linewidth]{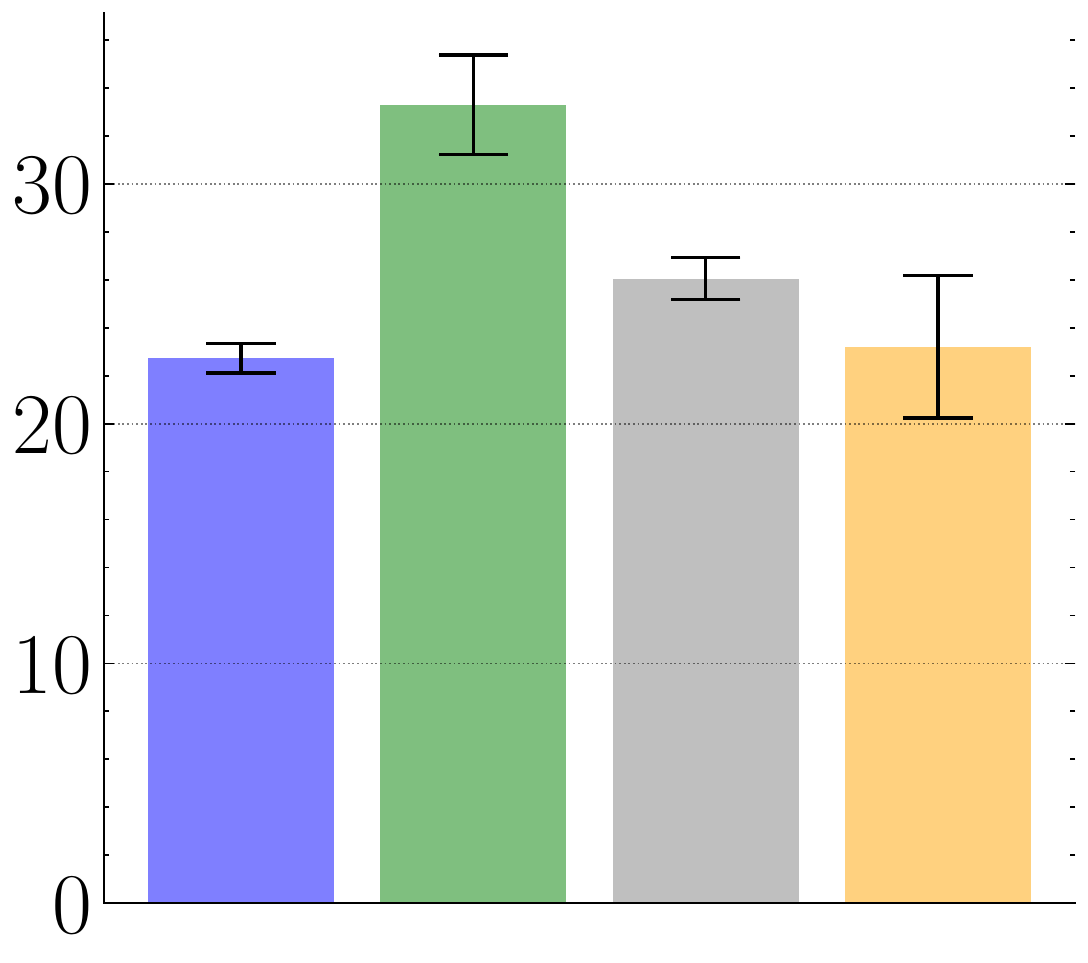} & \includegraphics[width=0.14\linewidth]{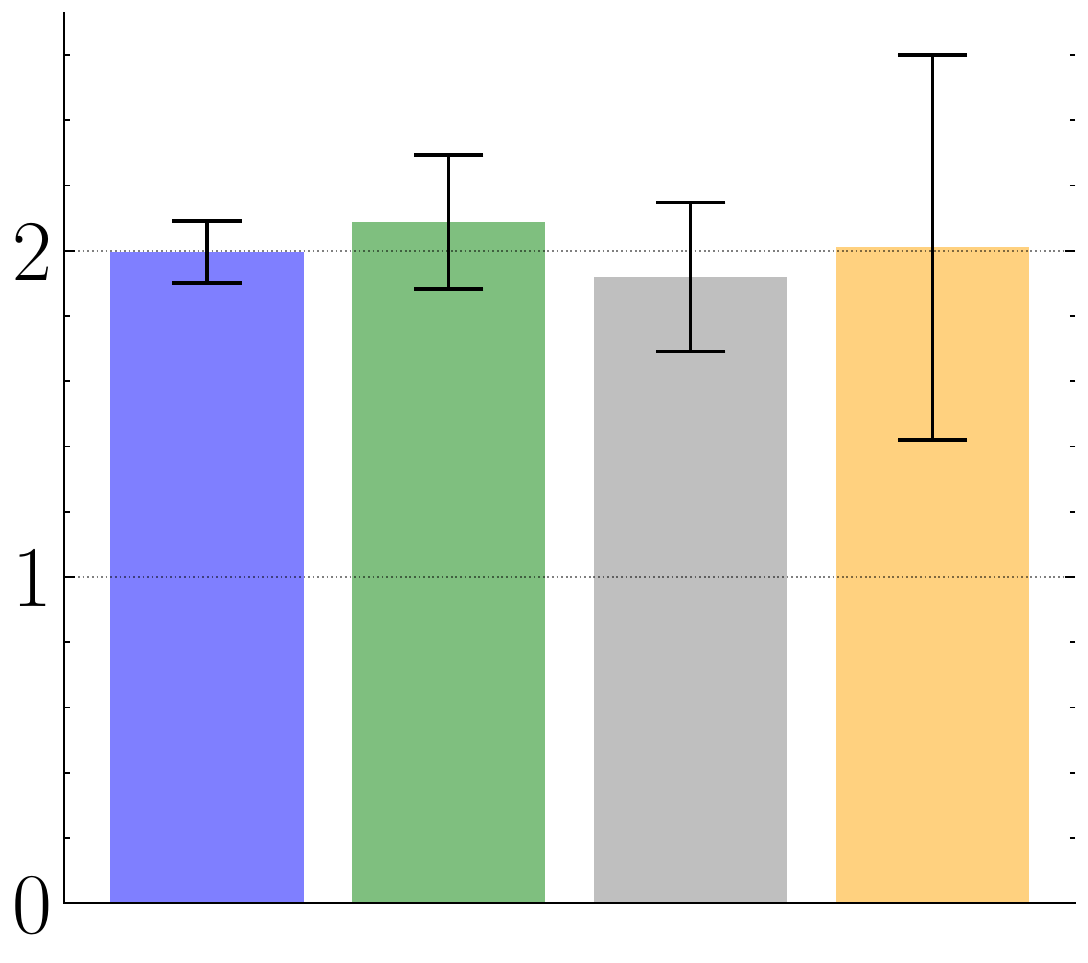} & \includegraphics[width=0.14\linewidth]{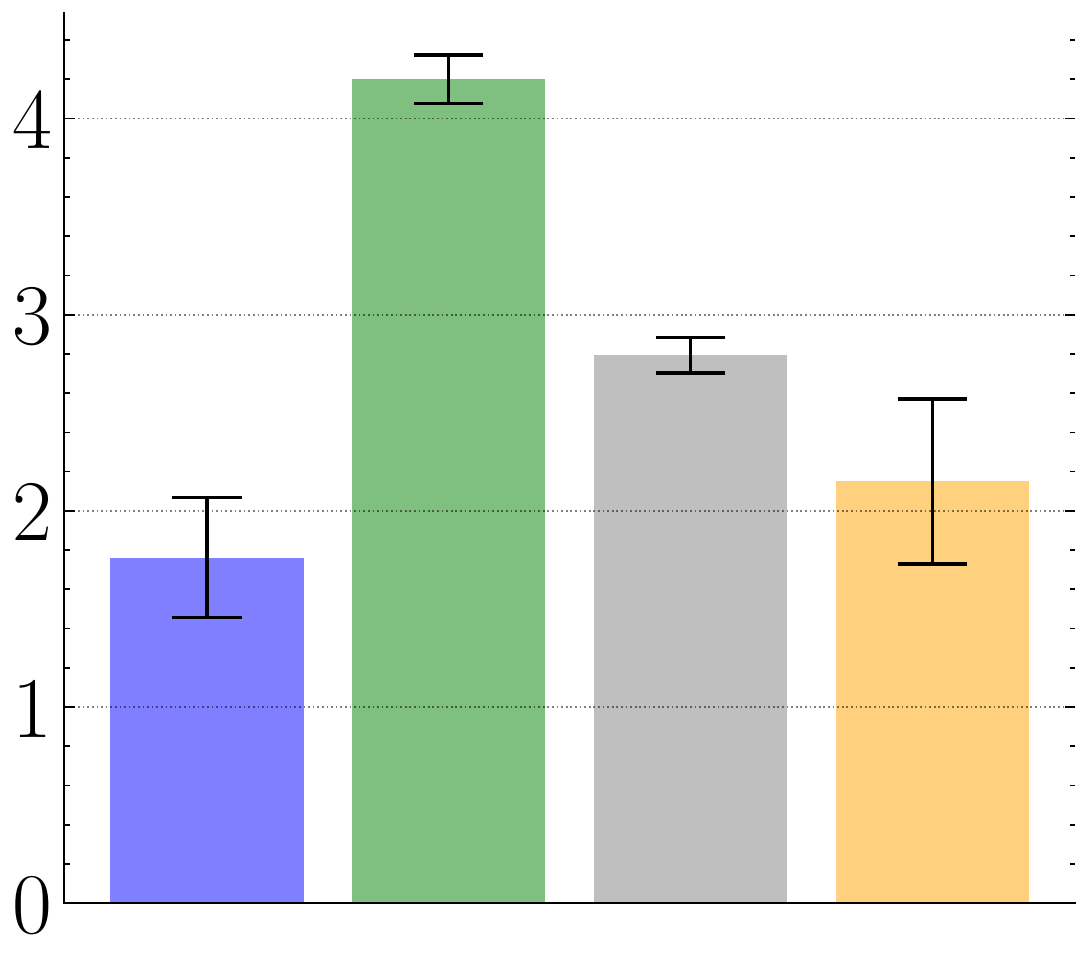} & \includegraphics[width=0.14\linewidth]{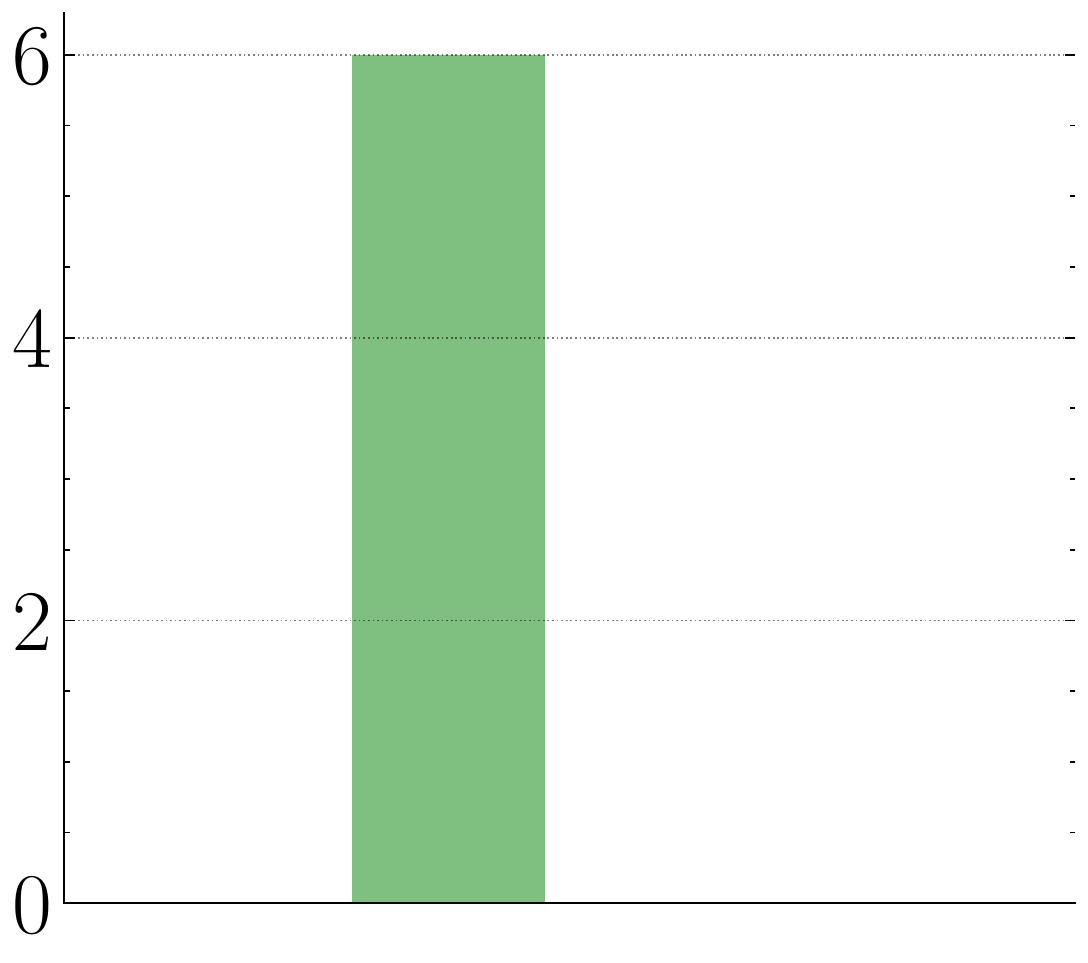}                                                                                        \\ \hline
    \rotatebox{90}{\parbox{0.14\linewidth}{\emph{Crossroad                                                                                                                                                                                                                                                                                                                                                                                                                              \\ Merge}}}     &
    \includegraphics[width=0.14\linewidth]{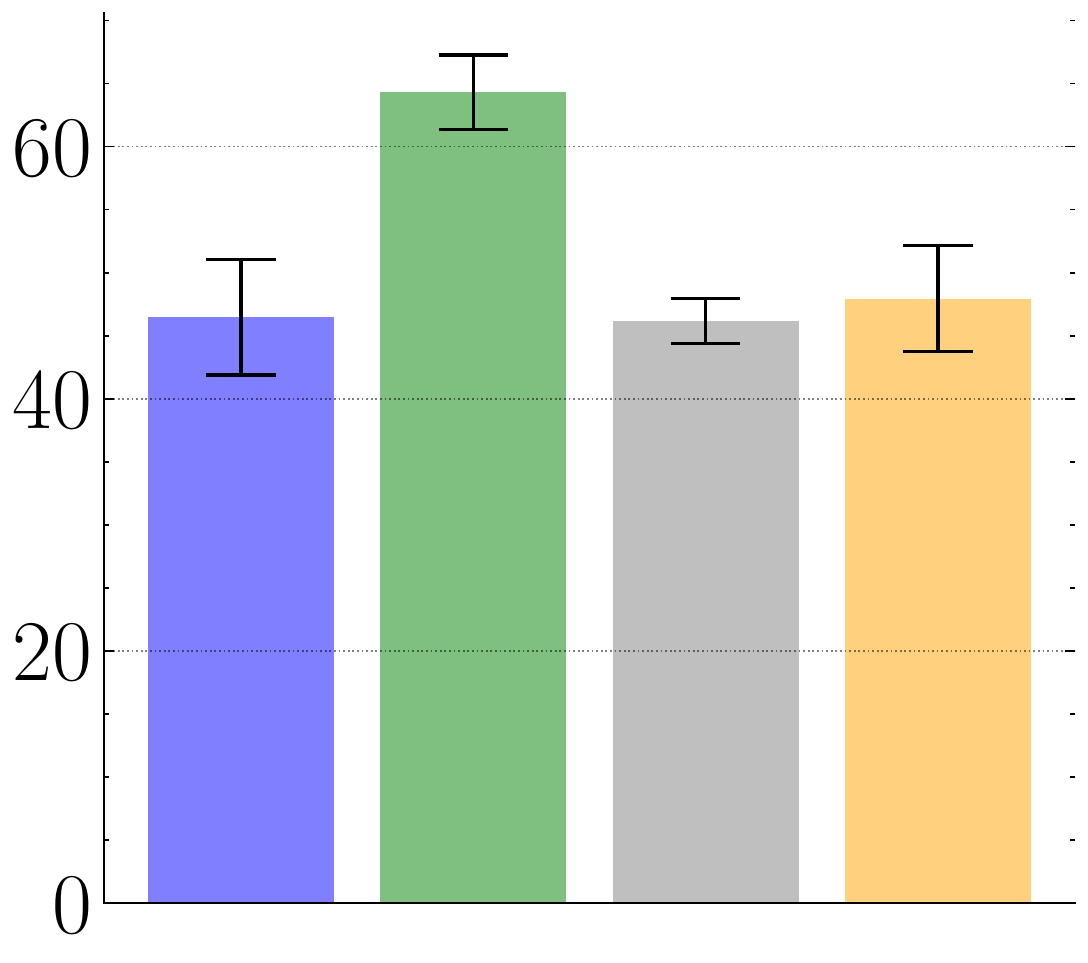}         & \includegraphics[width=0.14\linewidth]{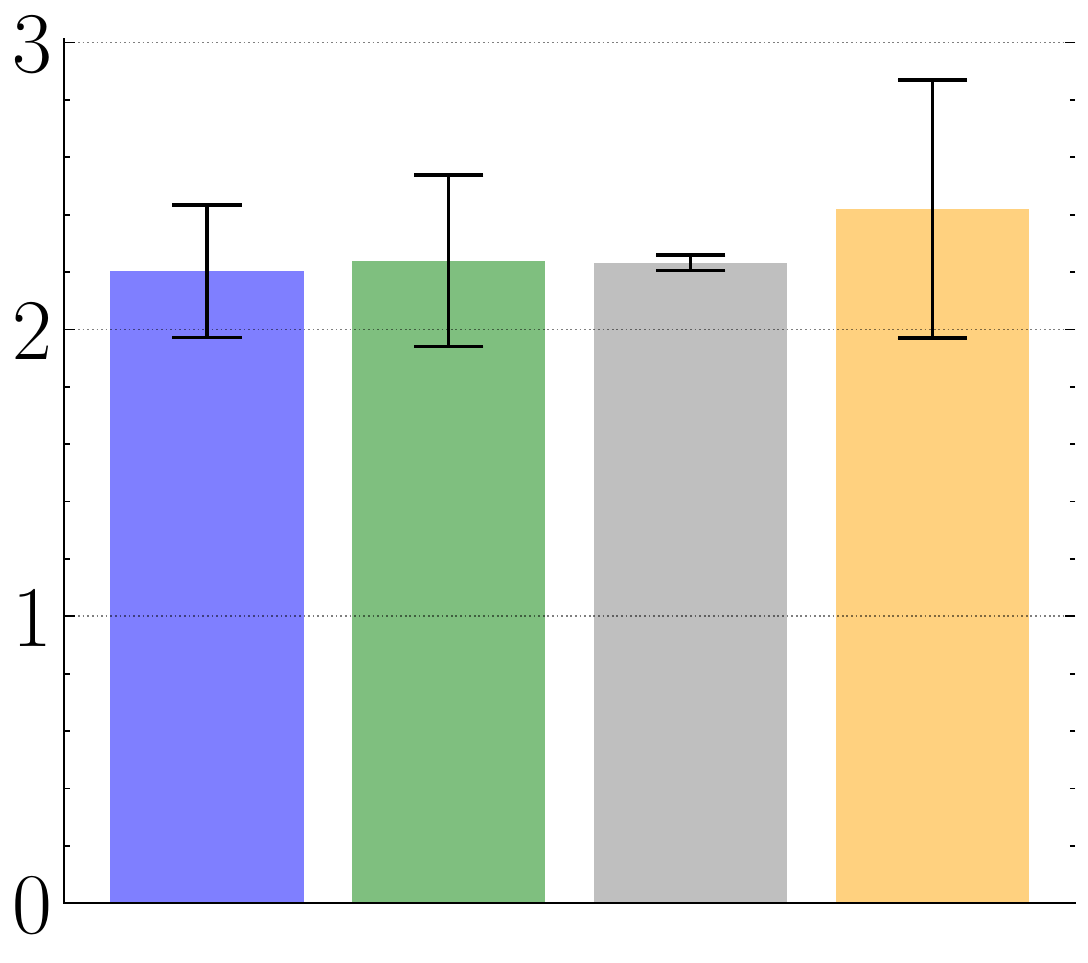}         & \includegraphics[width=0.14\linewidth]{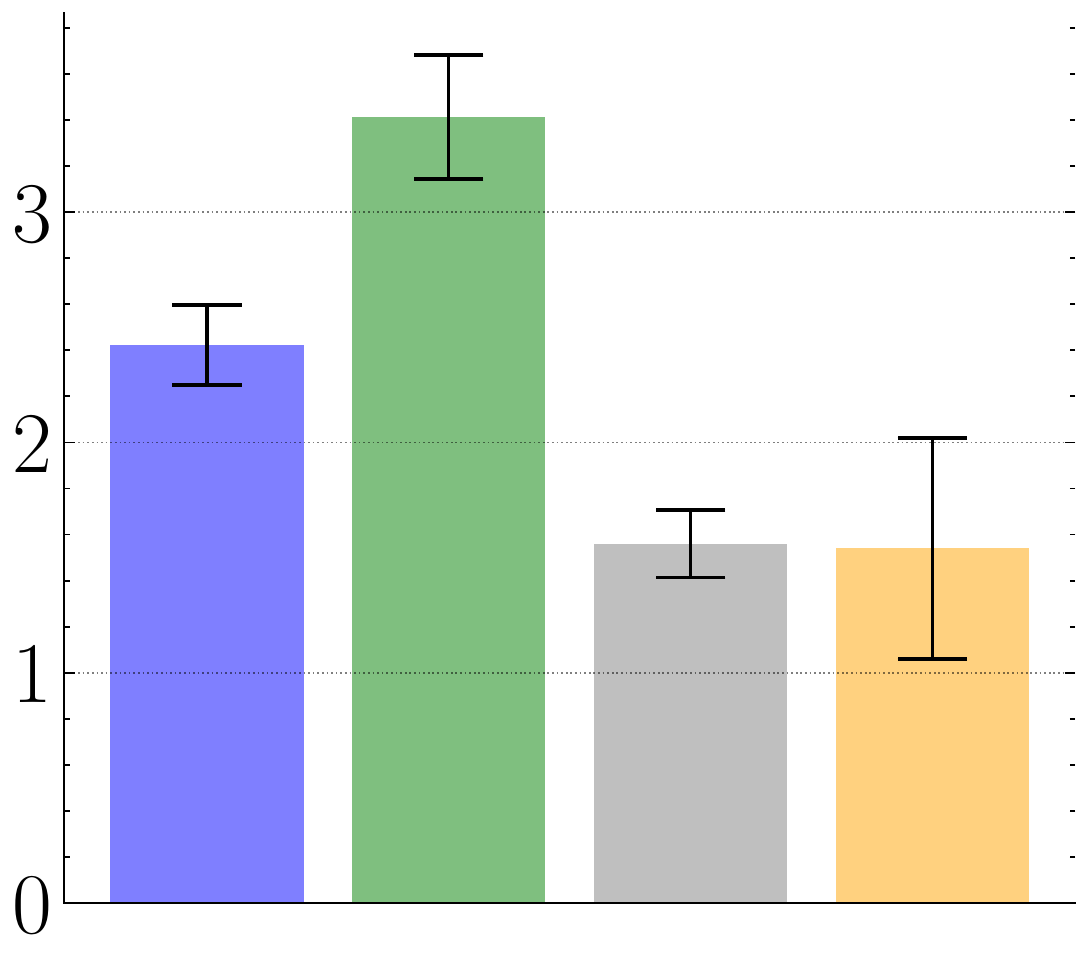}         & \includegraphics[width=0.14\linewidth]{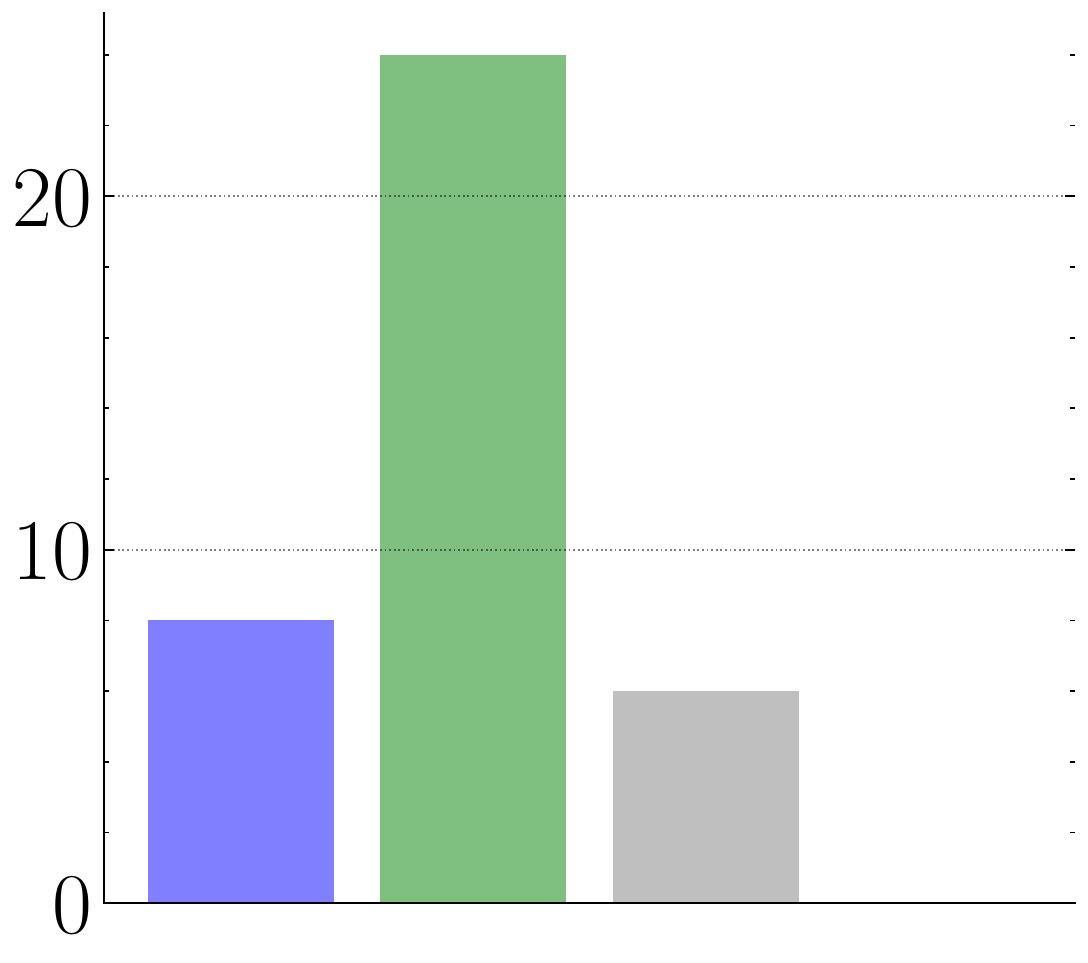}                                                                                                \\ \hline
    \rotatebox{90}{\parbox{0.14\linewidth}{\emph{Roundabout                                                                                                                                                                                                                                                                                                                                                                                                                             \\ Merge}}}             &
    \includegraphics[width=0.14\linewidth]{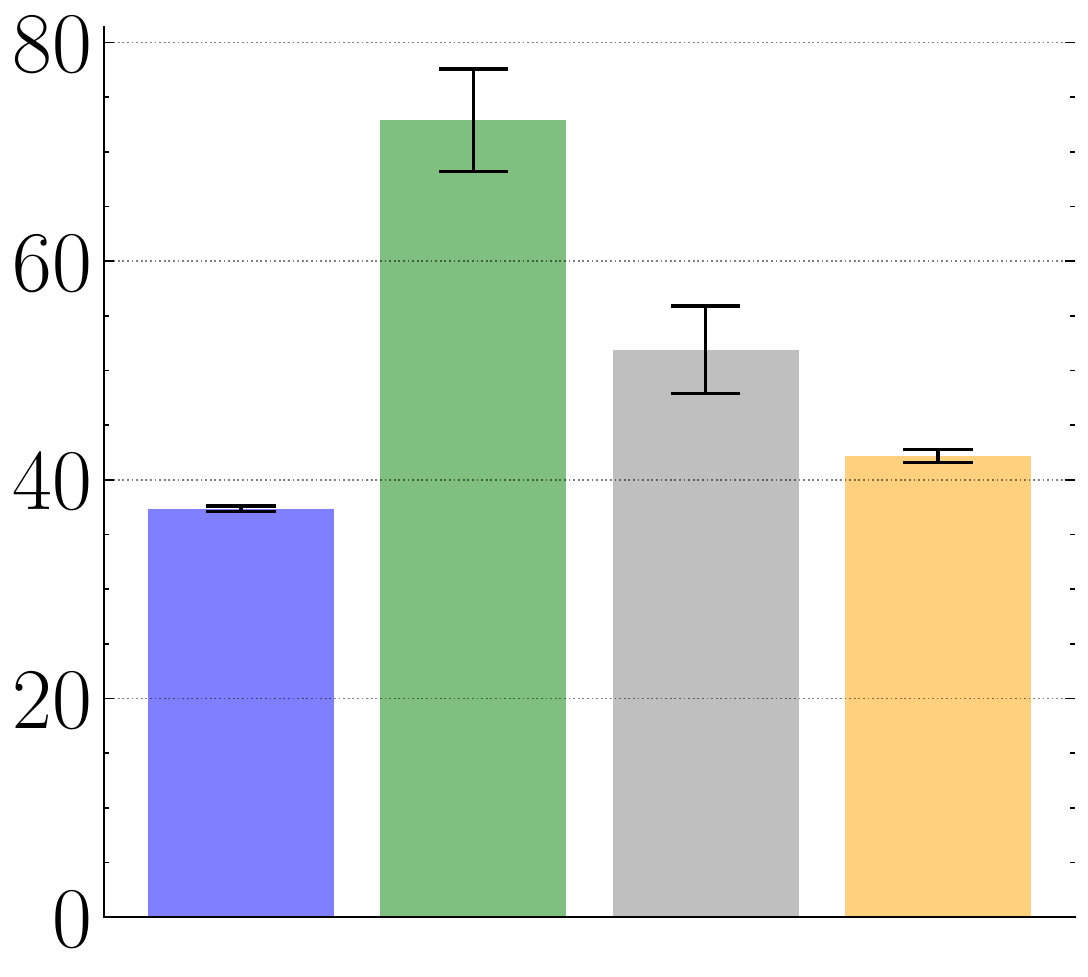}        & \includegraphics[width=0.14\linewidth]{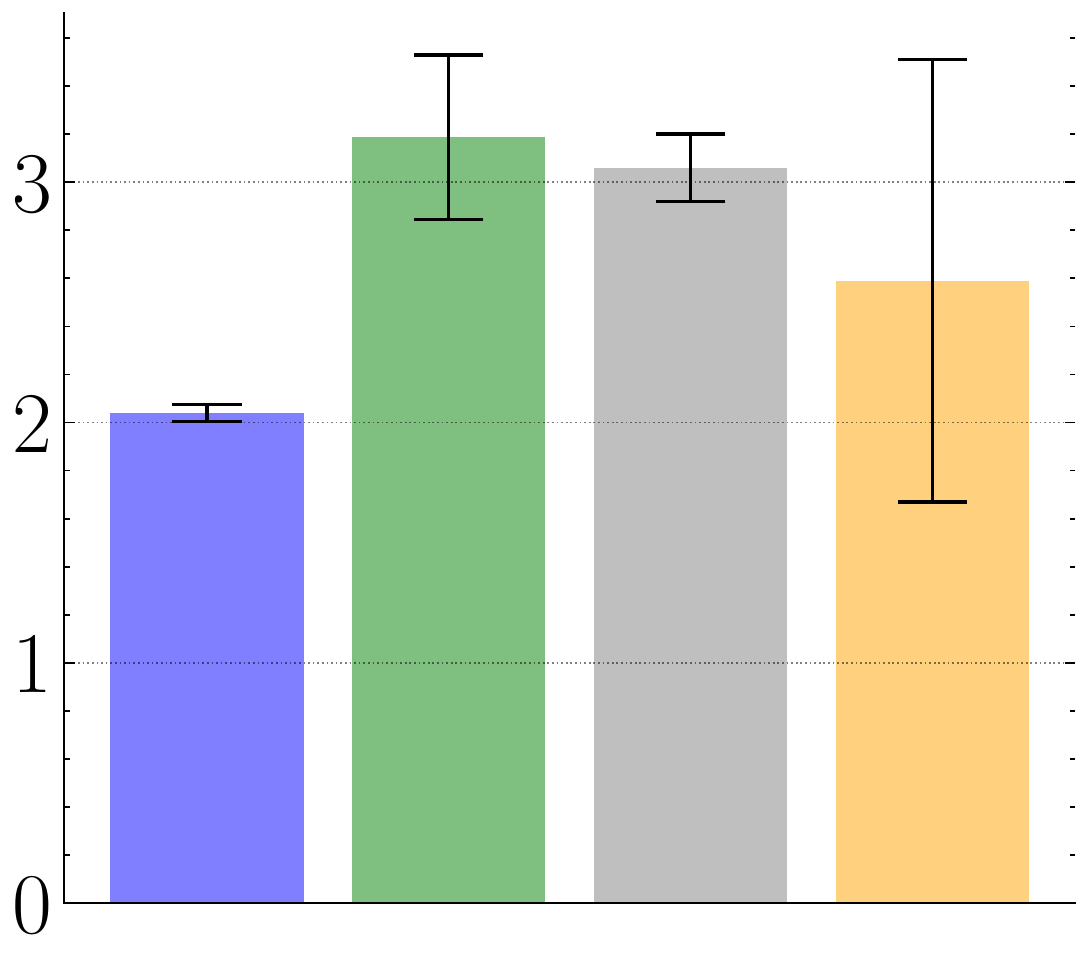}        & \includegraphics[width=0.14\linewidth]{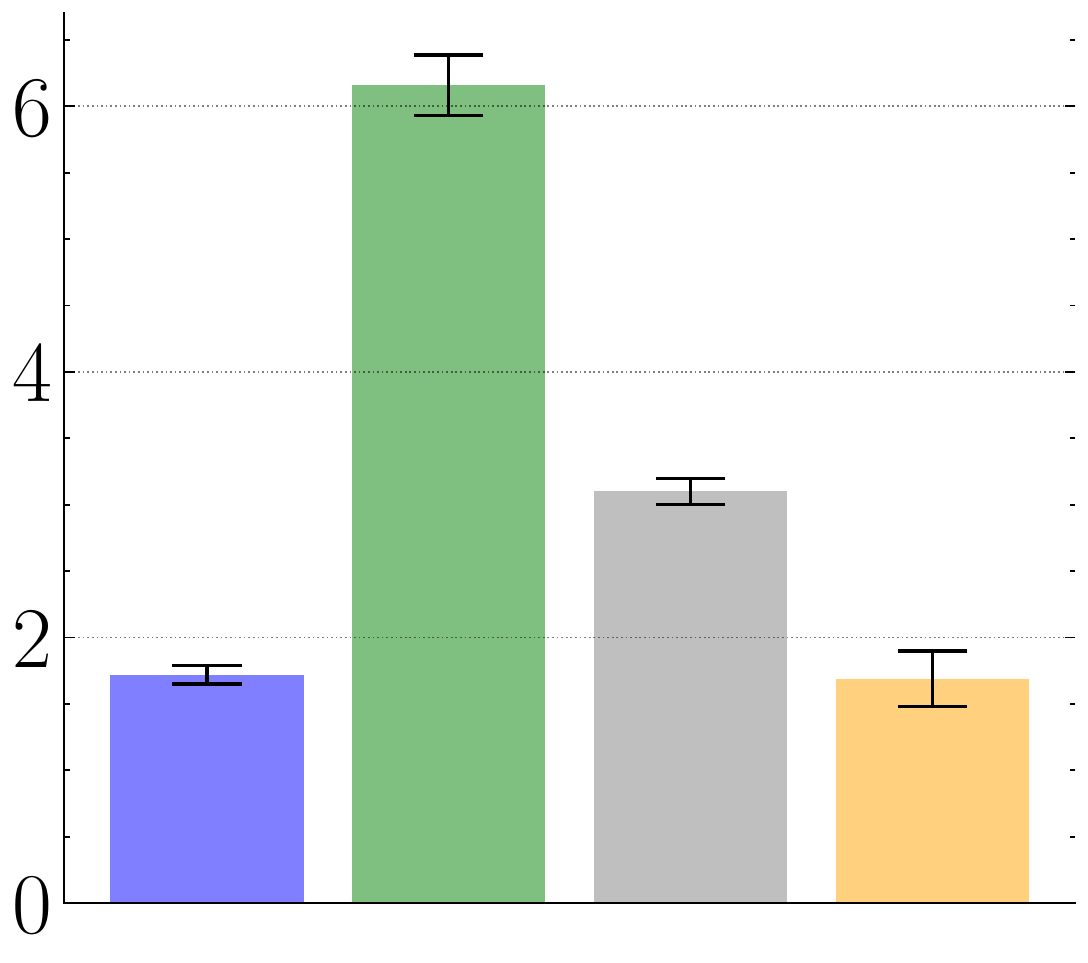}        & \includegraphics[width=0.14\linewidth]{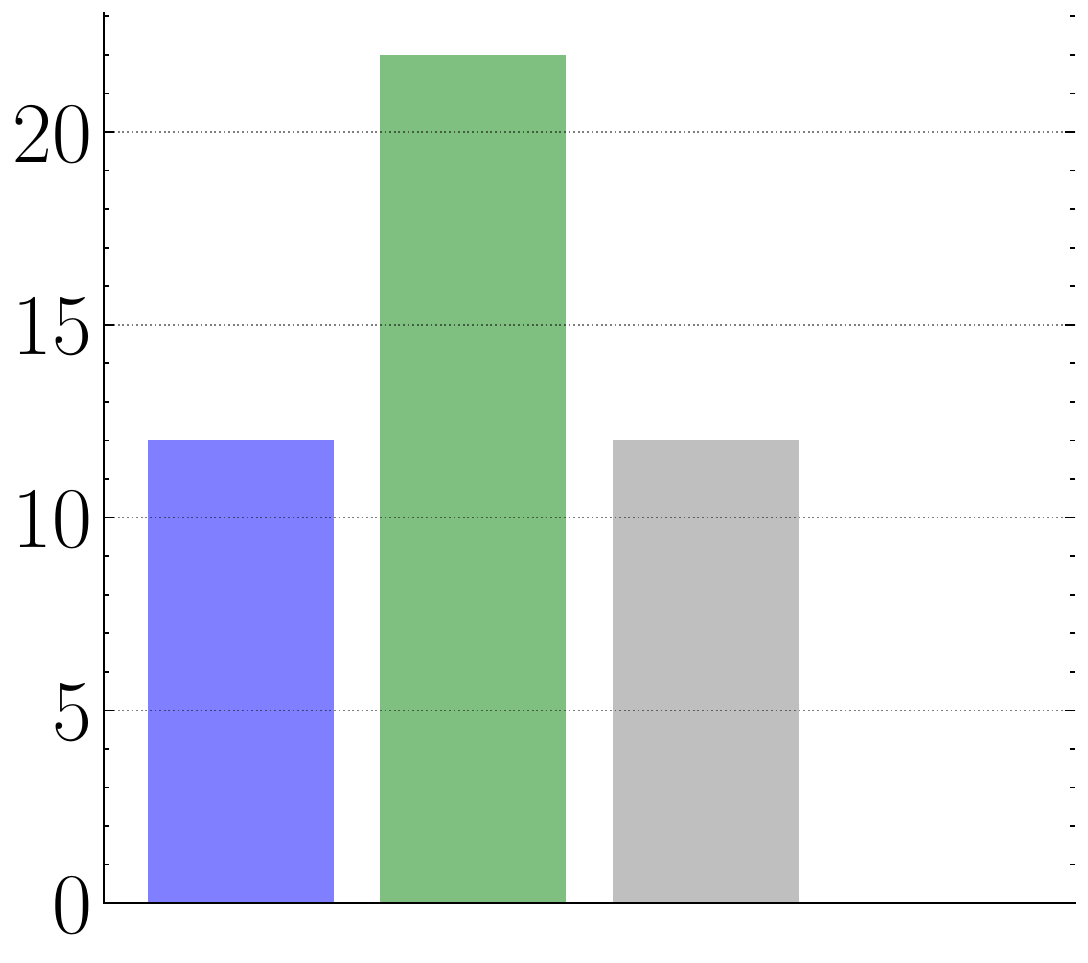}                                                                                               \\ \hline
    \rotatebox{90}{\parbox{0.14\linewidth}{\emph{Crossroad                                                                                                                                                                                                                                                                                                                                                                                                                              \\ Turn Left}}}        & \includegraphics[width=0.14\linewidth]{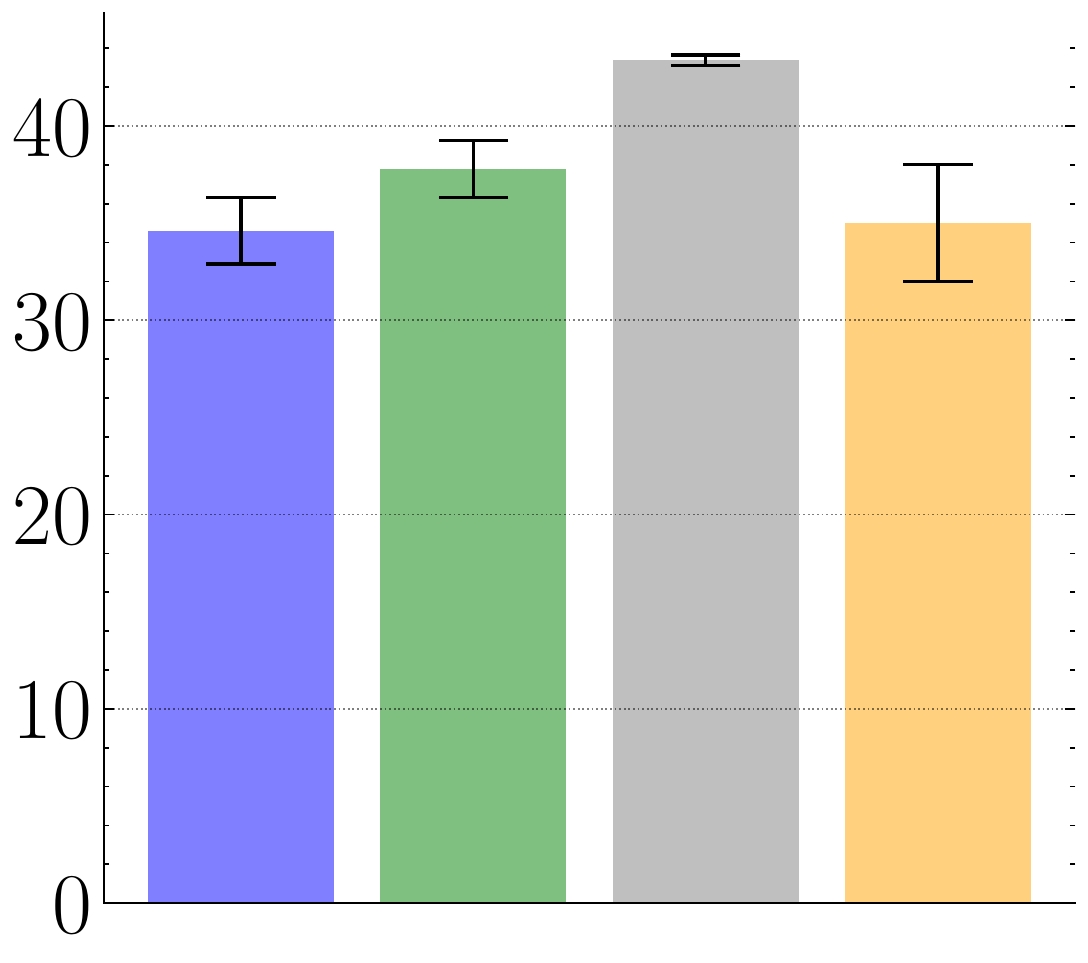}  & \includegraphics[width=0.14\linewidth]{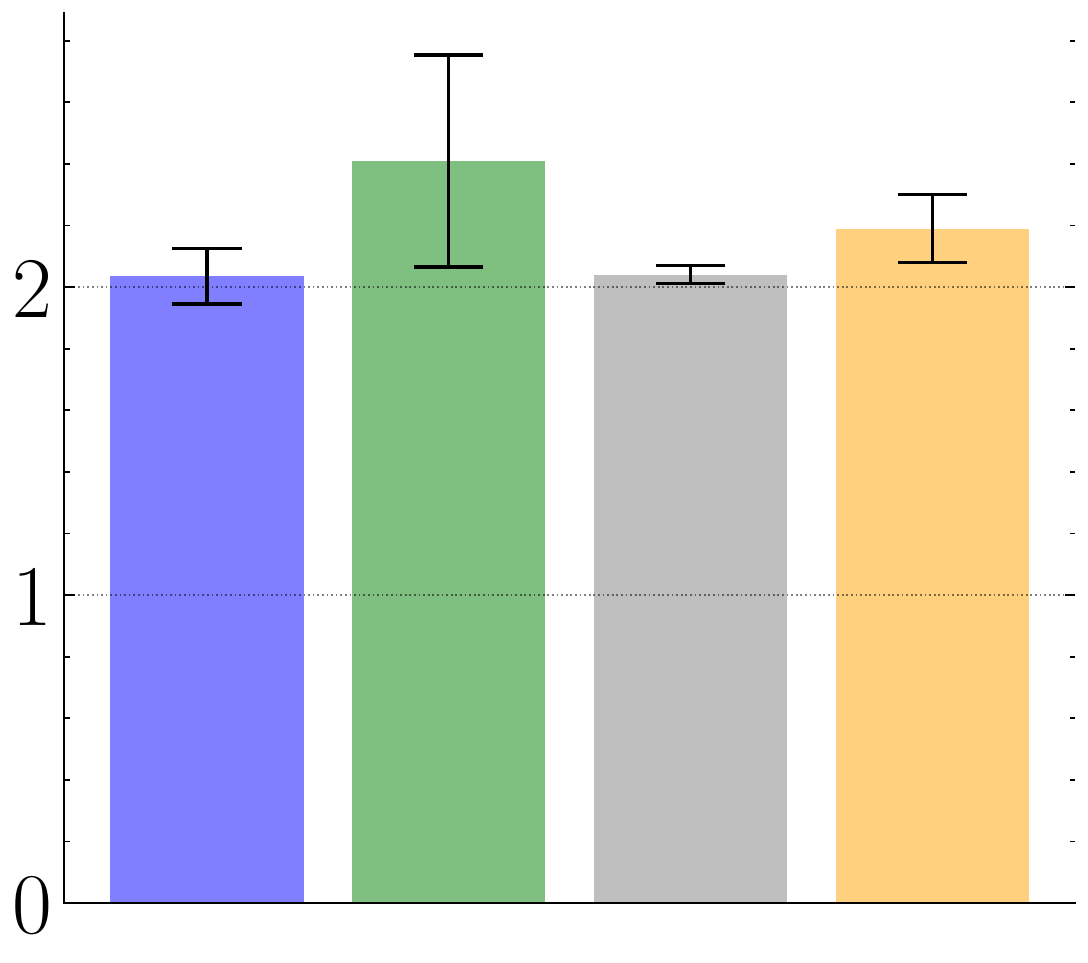} & \includegraphics[width=0.14\linewidth]{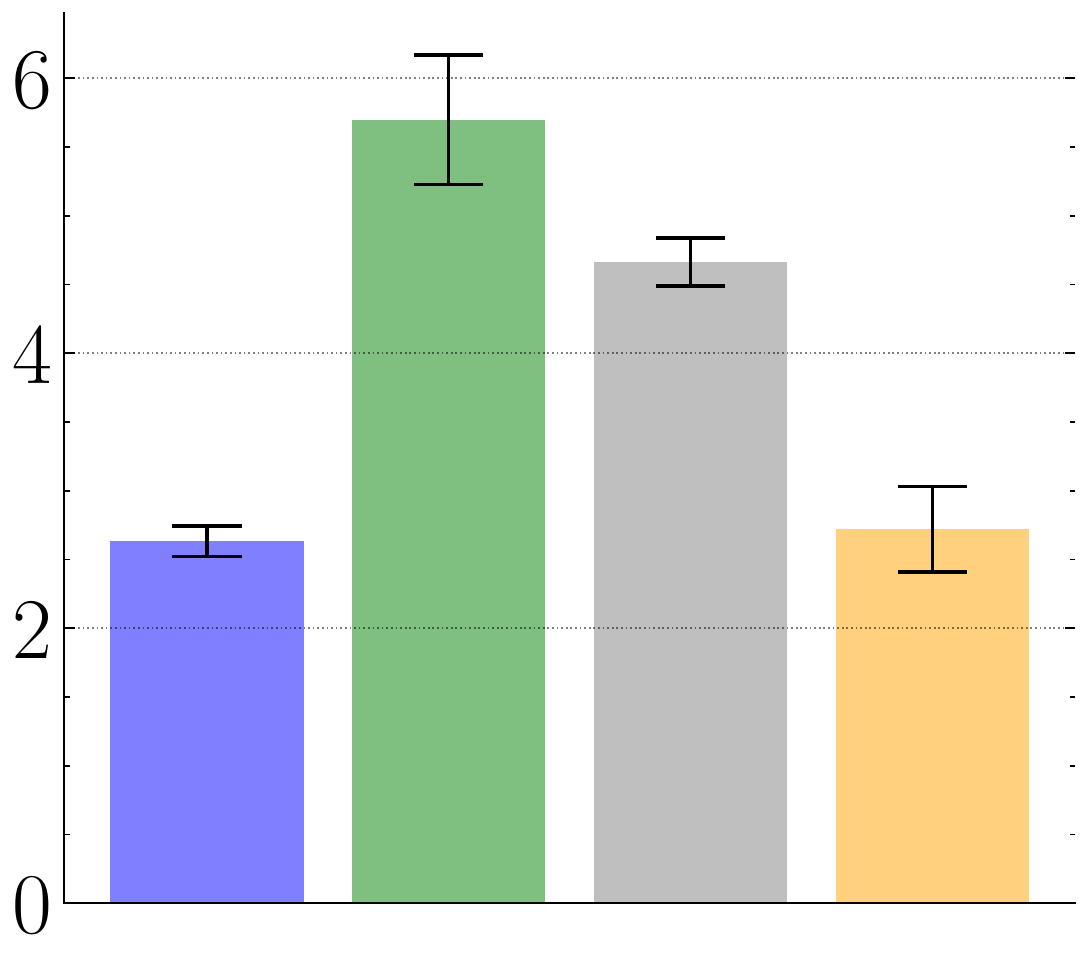} & \includegraphics[width=0.14\linewidth]{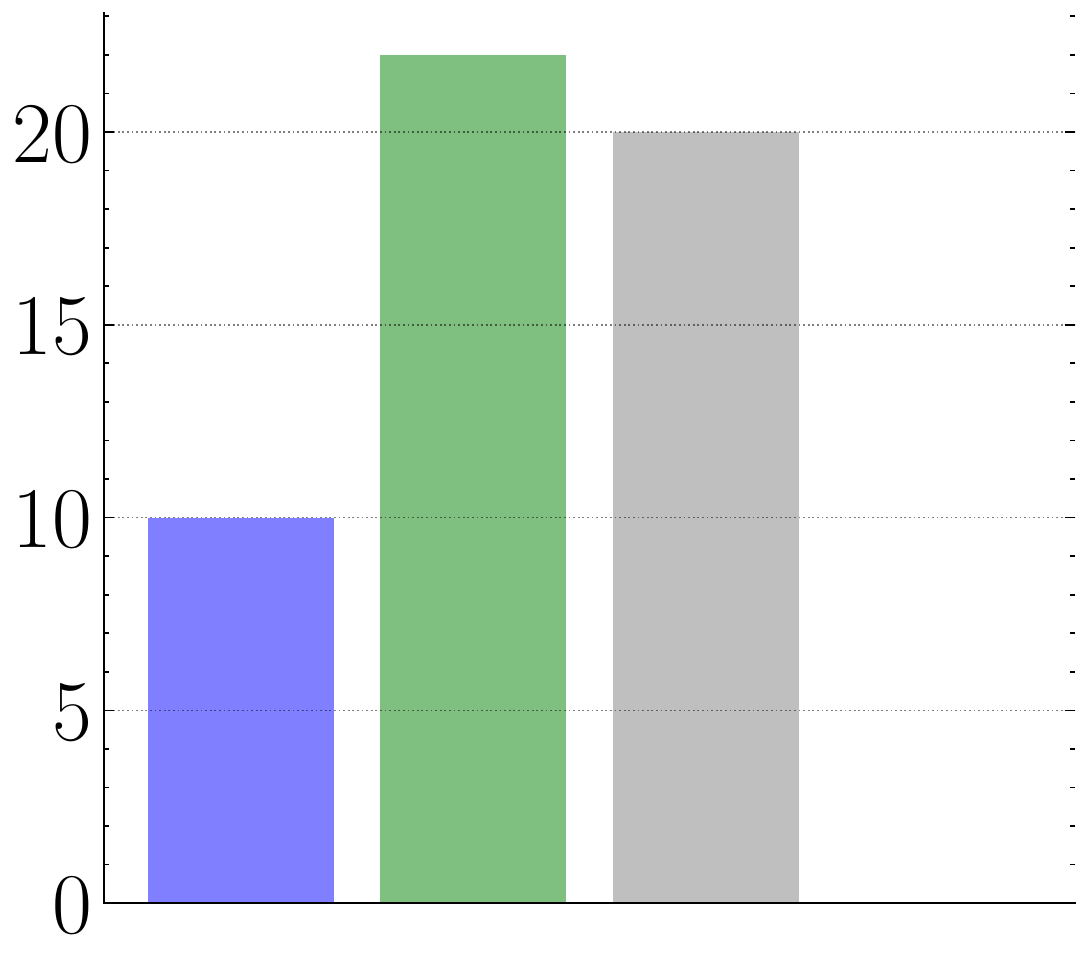}  \\ \hline
    \rotatebox{90}{\emph{Overtake}}                                                                & \includegraphics[width=0.14\linewidth]{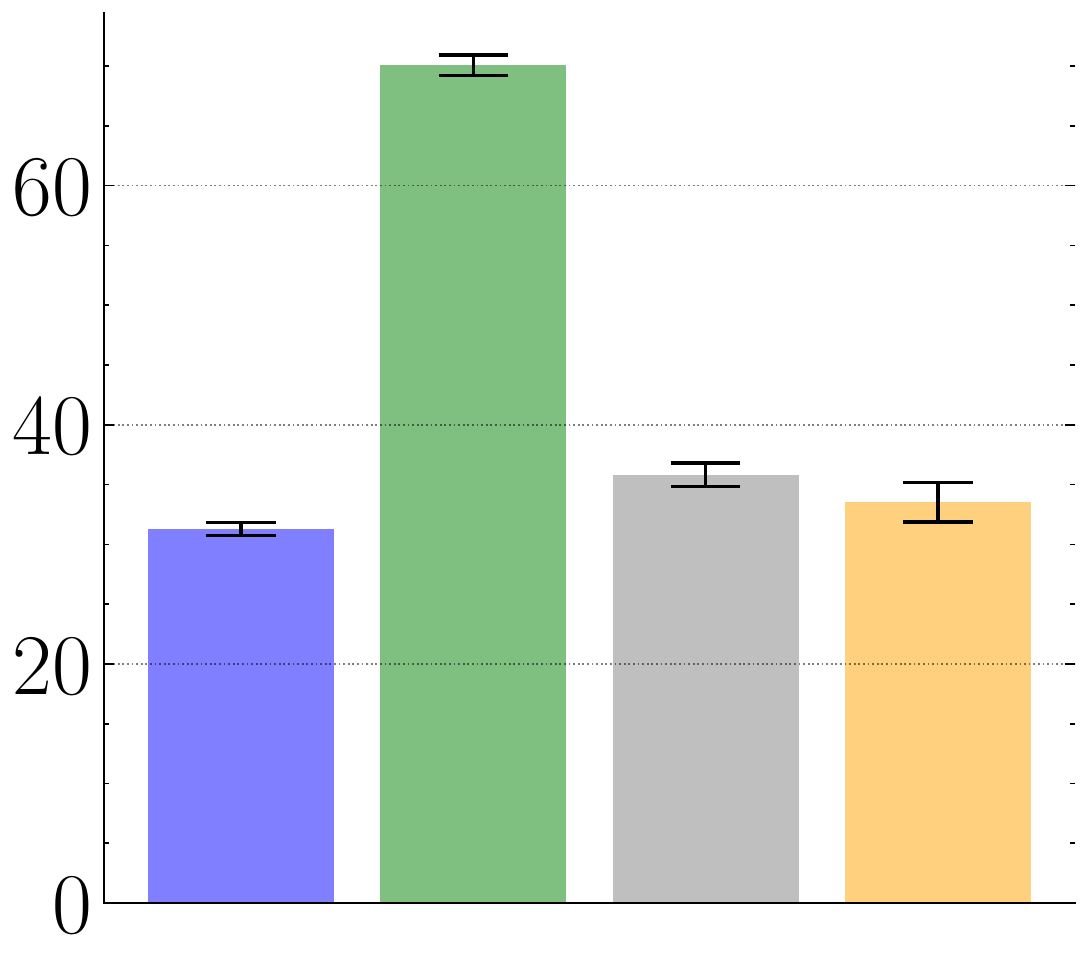}           & \includegraphics[width=0.14\linewidth]{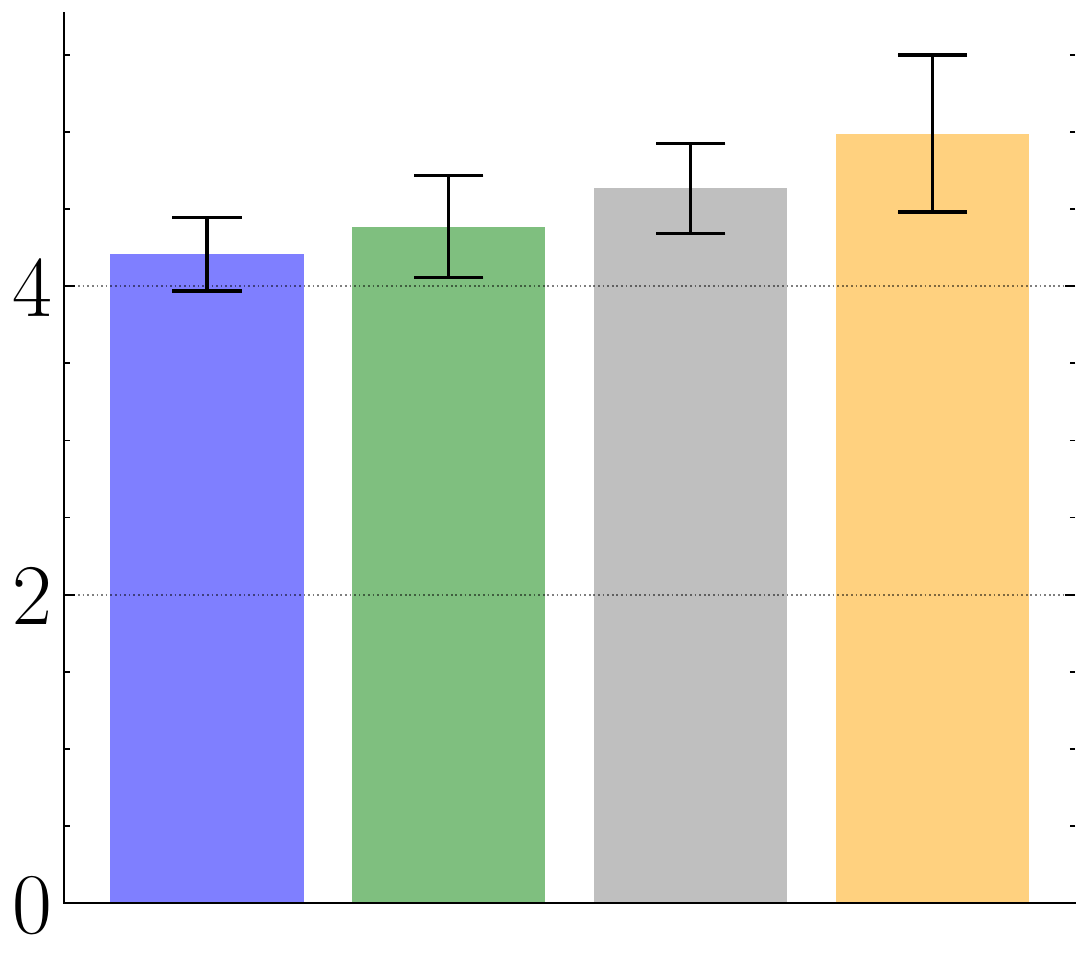}              & \includegraphics[width=0.14\linewidth]{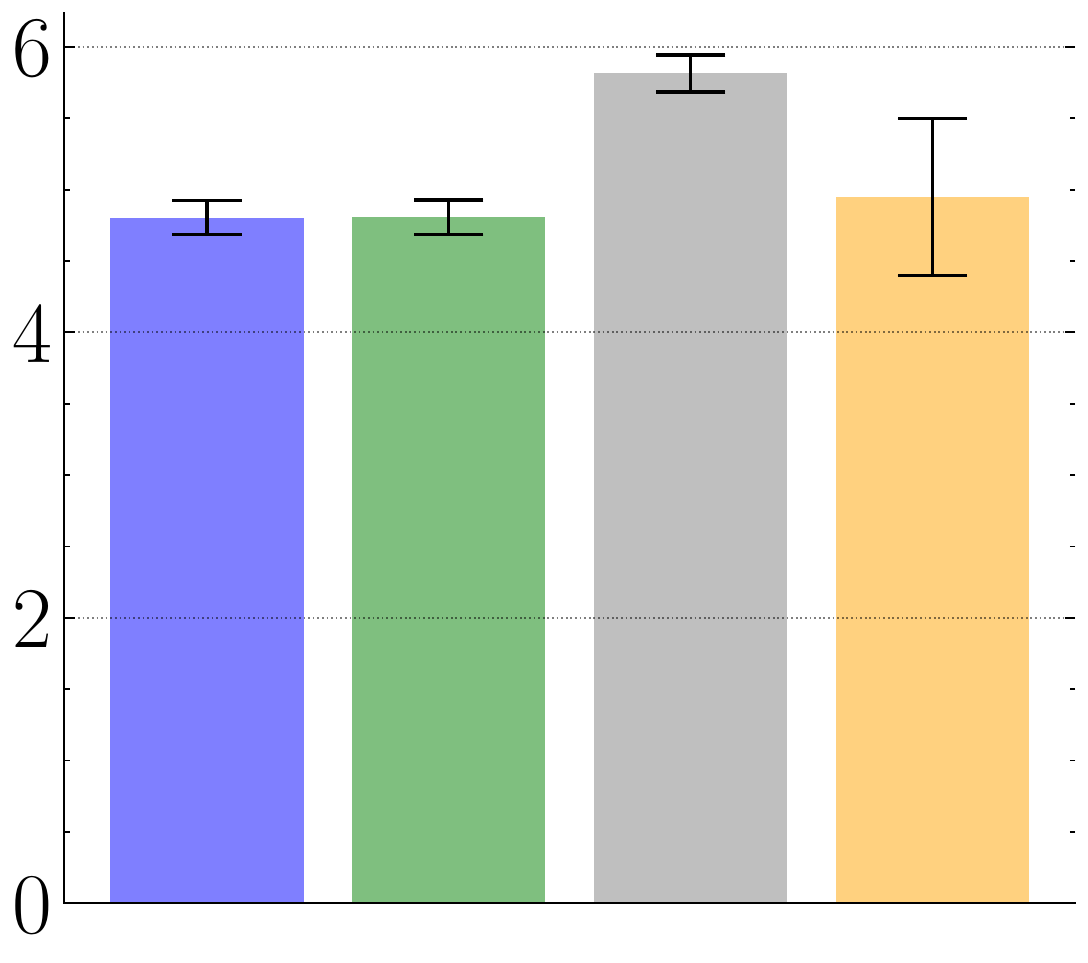}                        & \includegraphics[width=0.14\linewidth]{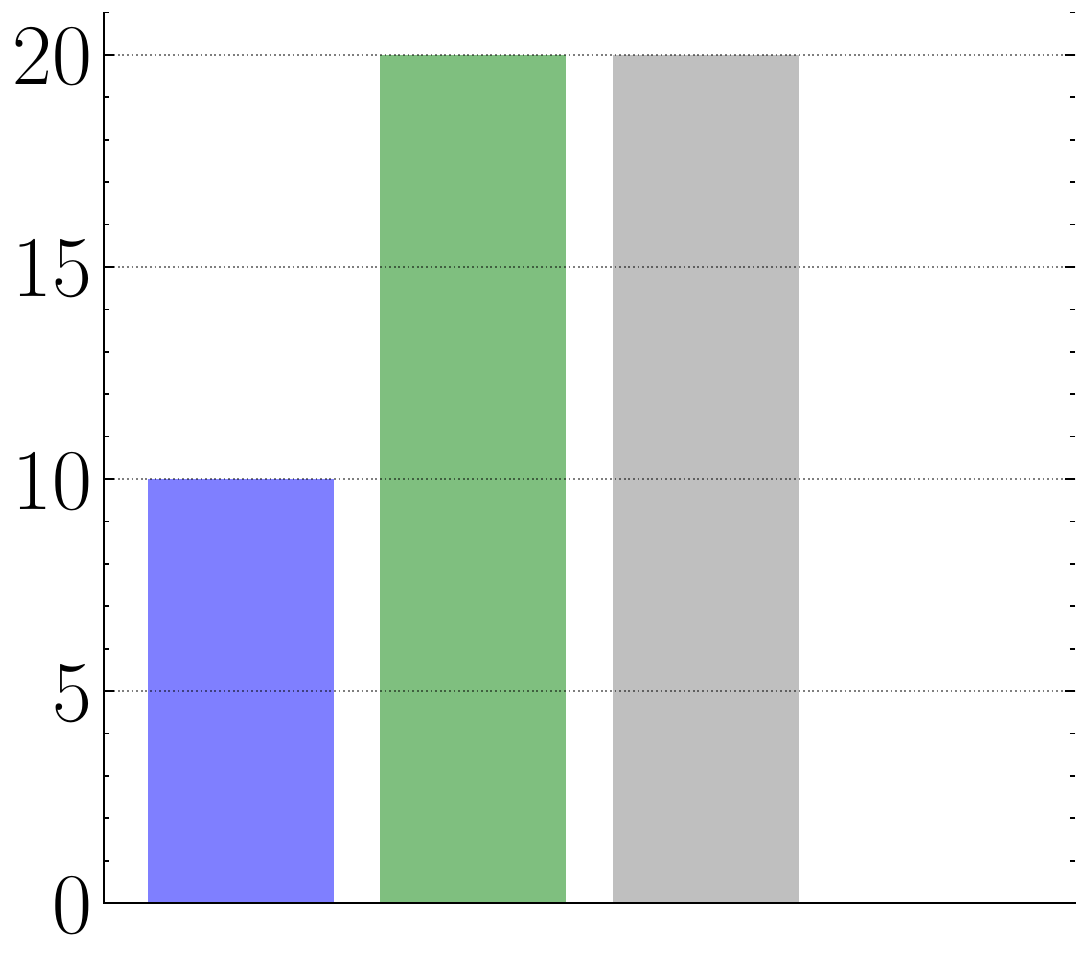} \\
    \bottomrule[1pt]
  \end{tabular}
  \vspace{-10pt}
\end{table*}
\vspace{-5pt}
\section{Conclusion}
\label{sec:conclusion}
\vspace{-5pt}
This work introduces a flexible framework for learning modular decision-making for autonomous driving. To learn the driver's high-level driving decision, the proposed framework imitates driver's control behavior with a modular generation pipeline. Being agnostic to the design of the non-differentiable downstream planning and control modules, this framework can train the neural decision module through reparametrized generative adversarial learning.  We evaluate its effectiveness in simulation environments with human driving demonstrations. This work can be extended to more complex environments where other vehicles are also learning agents, \textit{i.e.} multi-agent learning system. Alternatively, if we can collect data in the real-world and replay them in the simulator, this framework could be tested in real driving environments.

%===============================================================================

% The maximum paper length is 8 pages excluding references and acknowledgements, and 10 pages including references and acknowledgements

\clearpage
% The acknowledgments are automatically included only in the final version of the paper.
% \acknowledgments{If a paper is accepted, the final camera-ready version will (and probably should) include acknowledgments. All acknowledgments go at the end of the paper, including thanks to reviewers who gave useful comments, to colleagues who contributed to the ideas, and to funding agencies and corporate sponsors that provided financial support.}

%===============================================================================

\bibliography{example}  % .bib
% \iffalse
\newpage
\appendix
\section*{Appendices}
\addcontentsline{toc}{section}{Appendices}
\renewcommand{\thesection}{\Alph{section}}
\label{sec:appendix}
\section{Planning and Control Modules}
The planning module processes path and velocity separately. Paths are planned with cubic Bezier curve~\cite{5233184}:
\begin{equation}
  \begin{split}
    B(t)=(1-t)^3 P_s + 3(1-t)^2 t P_1
    +3(1-t)t^2 P_2 + t^3 P_g, \quad t\in [0,1] \\
  \end{split}
  \label{eq:bezier}
\end{equation}
where $P_s$, $P_g$ are the starting and goal point, $P_1$ and $P_2$ are intermediate control points, imposing the geometrical constraints such as end pose aligning with lane direction.

The velocity planning module reparameterizes Eq.(\ref{eq:bezier}) with arc-length, introducing the temporal unit. This reparameterization is not done directly, rather, $n$ spline segments with identical time unit $\Delta T$ are used to fit sample points from (\ref{eq:bezier}). Each of the segments is defined as:
\begin{equation}
  s_i=a_i+b_it+c_it^2+d_it^3+e_it^4,\ t\in[0,\Delta T].
\end{equation}
Here we consider the constraint on velocity planning, by minimizing the acceleration and jerk while fitting to the sample points from (\ref{eq:bezier}) and maintaining the continuity of curvature. Formally, the fitting is done by Quadratic Programming (QP):
\begin{equation}
  \begin{split}
    \min \sum_{i=1}^{n}[\int_{0}^{\Delta T} \ddot{s}_{i}^{2} dt&+\int_{0}^{\Delta T}\dddot{s}_{i}^{2} dt],\\
    s.t.\ \dot{s}_0(0)=v_{now}, \ddot{s}_0(0)&=a_{now}, \dot{s}_n(\Delta T)=v_{goal}\\
    \dot{s}_k(\Delta T)=\dot{s}_{k+1}(0), & \ddot{s}_k(\Delta T)=\ddot{s}_{k+1}(0), for\ k=0,1...n-1\\
  \end{split}
\end{equation}
% Details on solving this QP is discussed in the appendix. 

The control module is a basic PID controller. Here to reconcile with the MDP assumption in decision, we adopt the discrete-time form:
\begin{equation}
  \begin{split}
    u(t)=K_p e(t_k)+ K_i \sum_{i=0}^{k}e(t_k)\Delta t+K_d \frac{e(t_k)-e(t_{k-1})}{\Delta t},
  \end{split}
\end{equation}
where $K_p$, $K_i$ and $K_d$ denote the coefficients for the proportional, integral, and derivation terms respectively. $e(t)$ is the error function. In this work we have two independent error functions $e_{lat}(t)$ and $e_{lon}(t)$ for lateral controller steering wheel $u_{steer}(t)$ and longitudinal controller throttle/brake $u_{v}(t)$, which are both capped by mechanical constraints.
\section{Training Details}
\subsection{Experimental Setup}

\begin{figure}[htp]
  \centering
  \begin{minipage}{0.55\linewidth}
    \includegraphics[width=\linewidth]{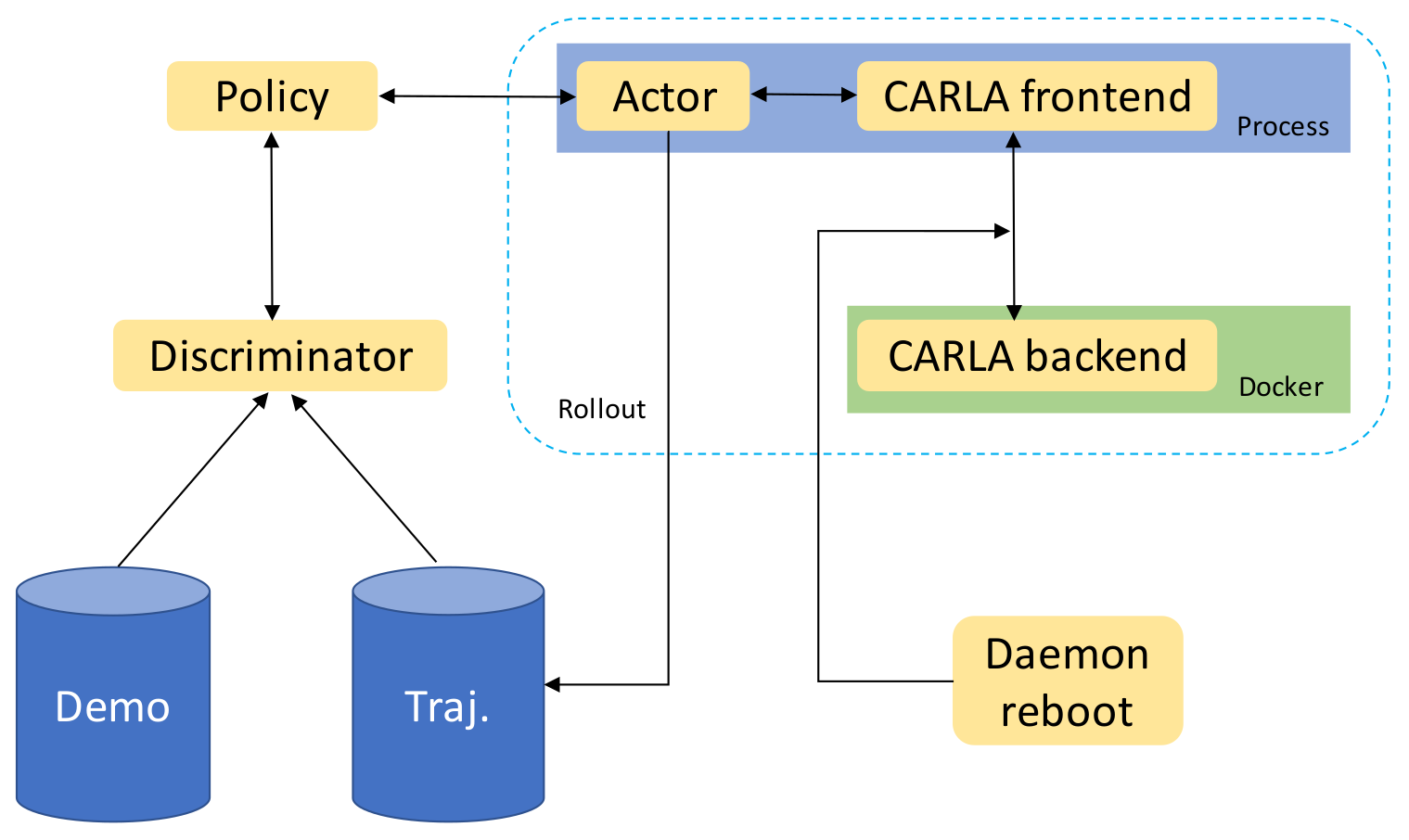}
    \caption{\textbf{Parallel Training System.} The first part(denoted as a blue dotted line box named Rollout) collects the transition data and the second part updates the model's parameters with the transition data. }
    \label{fig:parallel}
  \end{minipage}
  \hspace{15pt}
  \begin{minipage}{0.35\linewidth}
    \includegraphics[width=\linewidth]{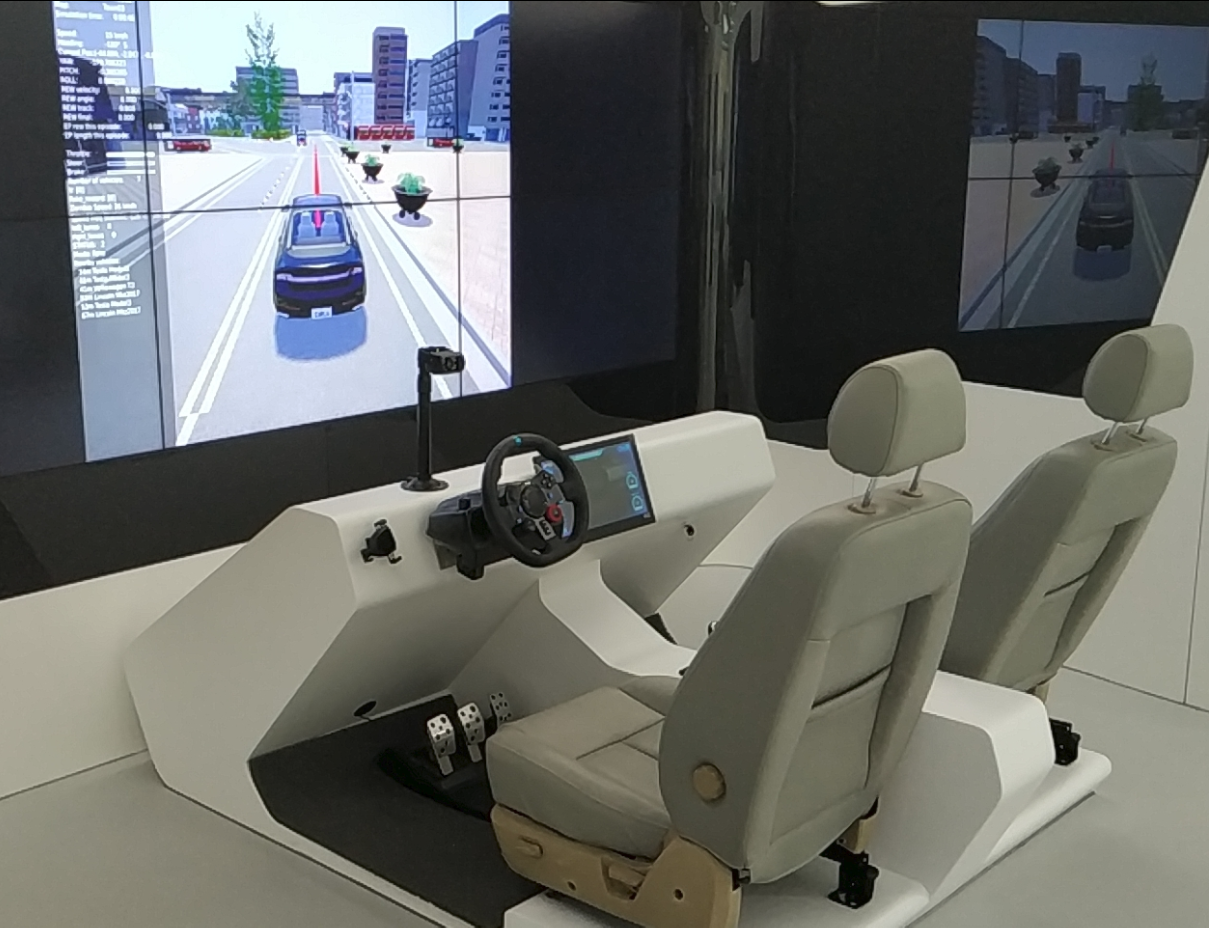}
    \caption{\textbf{Autonomous Driving Simulation Platform}, which is used to collect the low-level drivers' control behavior. We learn high-level driving decisions by imitating the low-level drivers' control commands.}% We collect the human demonstration data by driving a high-fidelity }
    \label{fig:real}
  \end{minipage}
\end{figure}

\paragraph{Training System} We evaluate the proposed framework in CARLA simulation platform \cite{Dosovitskiy17}. As the proposed framework focuses on the decision and its downstream, the state estimator is neglected and the observations are extracted directly from the simulator. Since the most recent version of CARLA is not efficient enough for large-scale simulation and learning, we build a parallel training system. The parallel training system shown in Fig.\ref{fig:parallel} consists of two parts. The first part(denoted as a blue dotted line box named Rollout) collects the transition data and the second part updates the model's parameters with the transition data. Every iteration, in Rollout, an actor synchronizes with the Policy node and obtains the model's parameters. Then the actor retrieves the transition data via the CARLA frontend. After that, the transition data would be stored in the Trajectory node. Here the transition data is generated by the interaction between the CARLA frontend and the CARLA backend in a Docker container. After the collection of the transition data finished, we start to update our model's parameters. The Discriminator and Policy nodes would update their parameters by following Algorithm \ref{alg:main} and using the data in the Trajectory and Demo nodes. To improve the robustness of the system, a Daemon reboot node is injected to reboot the Docker container when the system hangs.

% We build a parallel training system to improve training efficiency. 
%Since the most recent version of CARLA is not efficient enough for large-scale simulation and learning, we build a parallel training system. 
%\textcolor{red}{Details are shown in the supplementary.}  

\begin{figure}[htp]
  \centering
  \begin{subfigure}{.23\textwidth}
    \centering
    % include second image
    \includegraphics[width=.95\linewidth]{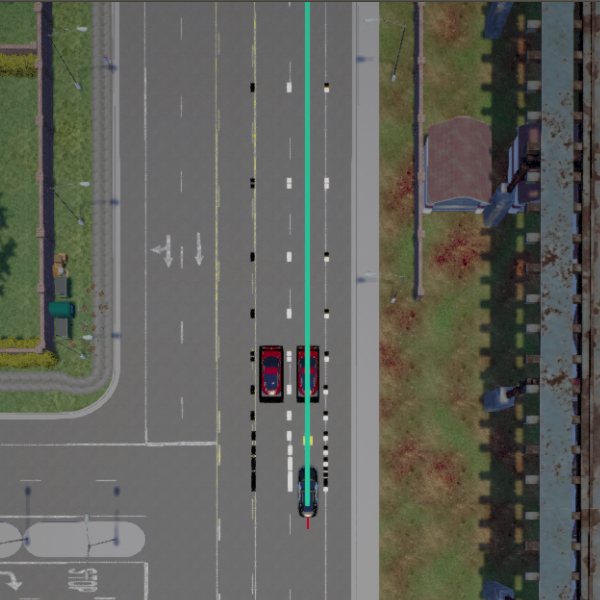}
    \caption{\scriptsize{Two-lane Car Following}}
    \label{subfig:two-lane-car-following}
  \end{subfigure}
  \begin{subfigure}{.23\textwidth}
    \centering
    % include first image
    \includegraphics[width=.95\linewidth]{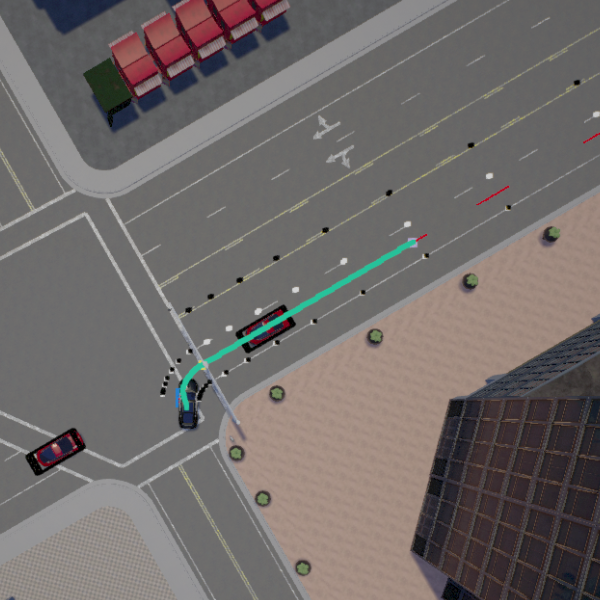}
    \caption{\scriptsize{Crossroad Merge}}
    \label{subfig:crossroad-merge}
  \end{subfigure}
  \begin{subfigure}{.23\textwidth}
    \centering
    % include first image
    \includegraphics[width=.95\linewidth]{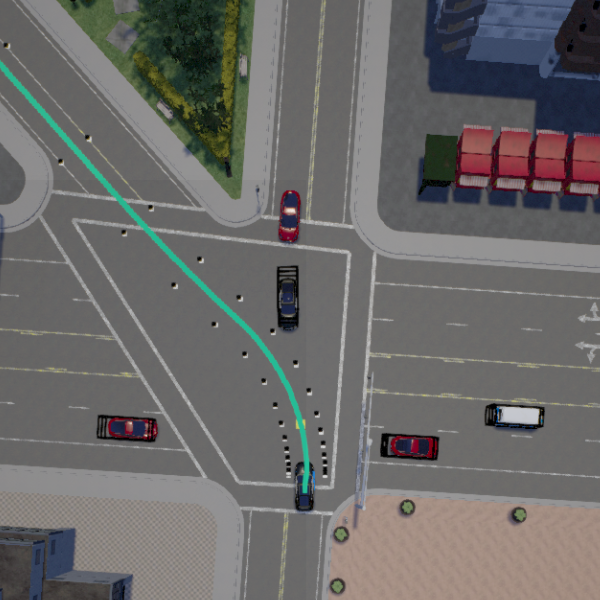}
    \caption{\scriptsize{Crossroad Turn Left}}
    \label{subfig:crossroad-turn-left}
  \end{subfigure}
  \begin{subfigure}{.23\textwidth}
    \centering
    % include first image
    \includegraphics[width=.95\linewidth]{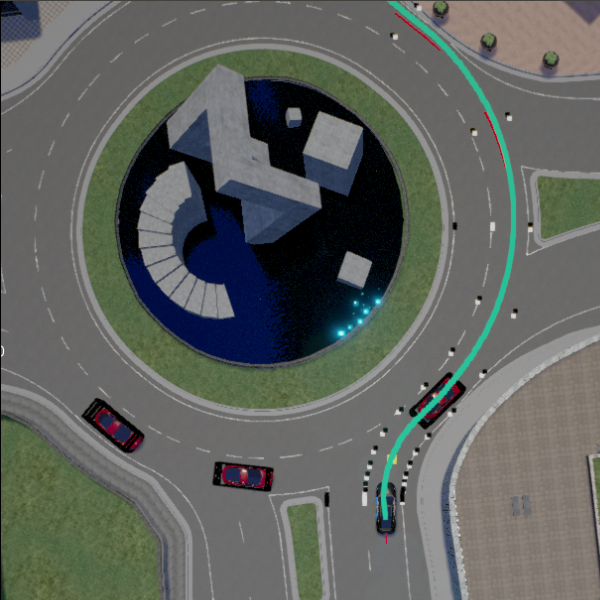}
    \caption{\scriptsize{Roundabout Merge}}
    \label{subfig:roundabout-merge}
  \end{subfigure}
  \caption{Screenshots of four scenarios.}
  \label{fig:scenarios}
\end{figure}

\paragraph{Scenarios Setup} Multiple common driving scenarios of varying complexity are constructed for evaluation. All driving scenarios are deployed in the Map Town 3 except Overtake. Since Town 3 provides only two-lane roads and Overtake should be demonstrated on a three-lane road, we choose Town 4 to design Overtake. To increase the complexity of the environment, the speed of the zombie car is set with randomness, and we give its average speed and design standard deviation.
\begin{itemize}
  \item \emph{Empty Town} is a special traffic scenario where there is no traffic. It covers urban road structures like a straight road, curve road, crossing, and roundabout.
  \item \emph{Car Following} (Fig.\ref{subfig:two-lane-car-following}). There are two sub-scenarios in this scenario: \emph{Two Lanes Car Following} (Fig.\ref{subfig:two-lane-car-following}) and \emph{Single Lane Car Following}.
        \emph{Two Lanes Car Following} is designed based on a four-lane dual carriageway. Two zombie cars are running side by side in two lanes, and the ego-car follows them in one of the lanes. The speed of the zombie cars is $18\pm 1$ km/h. In \emph{Single Lane Car Following}, only one zombie car is running ahead in a single lane, and the ego-car follows the zombie car. The speed of the zombie car is $15\pm 1$ km/h.
  \item \emph{Crossroad Merge} (Fig.\ref{subfig:crossroad-merge}). In the lateral lane, no less than three zombie cars as a traffic stream drive through the intersection in series, and ego-car merges into their traffic flow from a lane perpendicular to the lateral lane. The speed of zombie cars is randomly chosen from $21\pm 1$ km/h, $25\pm 1$ km/h. There are two sub-scenarios.
  \item \emph{Crossroad Turn Left} (Fig.\ref{subfig:crossroad-turn-left}).  Multiple zombie cars drive in the lateral lane and the opposite lane. The ego-car needs to cross the intersection and then head to the left lane, during which it needs to avoid collision with the zombie cars and obey the traffic rules. The speed of zombie cars is randomly chosen from $26\pm 1$ km/h, $42\pm 1$ km/h. There are two sub-scenarios.
  \item \emph{Roundabout Merge} (Fig.\ref{subfig:roundabout-merge}). Four zombie cars are entering the roundabout one by one, bypassing $270$ degrees counterclockwise and then driving out of the roundabout. The ego-car enters the roundabout and joins the traffic flow at about $90$ degrees. After following the zombie cars around the roundabout for about $180$ degrees, the ego-car drives out with them together. The speed of zombie cars is randomly chosen from $21\pm 1$ km/h, $25\pm 1$ km/h, $35\pm 1$ km/h. There are three sub-scenarios.
  \item \emph{Overtake} (Fig.\ref{fig:decision}). Ego Car drives in the middle lane (medium speed lane) of the three lanes with Zombie Car \uppercase\expandafter{\romannumeral 1\relax} in front of it.  Zombie Car \uppercase\expandafter{\romannumeral 2\relax} (different from Zombie Car \uppercase\expandafter{\romannumeral 1\relax}), which is on the left front of Zombie Car \uppercase\expandafter{\romannumeral 1\relax},  drives in the high-speed lane. Under the premise of obeying traffic rules, Ego Car has to go beyond  Zombie Car \uppercase\expandafter{\romannumeral 1\relax} in front of it by changing to the high-speed lane. Due to the blocking of Zombie Car \uppercase\expandafter{\romannumeral 2\relax}, Ego Car will change back to the medium speed lane after exceeding Zombie Car I, then completing the entire overtaking process.
\end{itemize}

The maneuvers of zombies in all scenarios are defined by two items: configurations of zombies, and a CARLA built-in waypoint-following controller. Configurations of zombies for experiments in Table \ref{tab:quant-comparison} is the footnote part in Table \ref{tab:scenario-configuration}.

Besides the car velocity variance mentioned above, the evaluation environments for each scenario also vary in seeds and vehicle models. For example, Audi TT, Chevrolet, Dodge (the police car), Etron, Lincoln, Mustang, Tesla 3S provided by CARLA.
\begin{table*}[htp]
  \centering
  \begin{threeparttable}
    \caption{Configurations of Scenarios}
    \label{tab:scenario-configuration}
    \begin{tabular}{>{\centering\arraybackslash}m{0.18\linewidth}>{\centering\arraybackslash}m{0.11\linewidth}>{\centering\arraybackslash}>{\centering\arraybackslash}m{0.09\linewidth}>{\centering\arraybackslash}m{0.09\linewidth}>{\centering\arraybackslash}m{0.17\linewidth}>{\centering\arraybackslash}m{0.09\linewidth}>{\centering\arraybackslash}m{0.09\linewidth}}
      \toprule[1pt]
      % \multirow{2}{*}{\textbf{Scenarios}}   & \multicolumn{4}{c}{\textbf{Learning-based Module}} & \textbf{Expert Data} \\
      \textbf{Scenarios}               & \textbf{Number of Cars} & \textbf{Starting Velocity of Ego Car ($km/h$)} & \textbf{Starting Velocity of Zombie Cars ($km/h$)} & \textbf{Target Velocity of Zombie Cars ($km/h$)}                                                                                                                                       & \textbf{Starting Direction ($^\circ$)} & \textbf{Expected Direction ($^\circ$)} \\ \hline
      % \cmidrule(lr){2-5}  \cmidrule(lr){6-6} 
      \toprule[1pt]
      \emph{Single Lane Car Following} & $2$                     & $25.20$                                        & $10.00$                                            & $15.00\pm 1.00$\footnotemark[3]                                                                                                                                                        & $0.00$                                 & $180.00$                               \\ \hline
      \emph{Two Lanes Car Following}   & $3$                     & $32.40$                                        & $11.00$                                            & $18.00\pm 1.00$\footnotemark[3]                                                                                                                                                        & $0.00$                                 & $0.00$                                 \\ \hline
      \emph{Crossroad Merge}           & $ > 4$                  & $25.20$                                        & $10.00$                                            & $21.00\pm 1.00$\footnotemark[3] $25.00\pm 1.00$                                                                                                                                        & $0.00$                                 & $90.00$                                \\ \hline
      \emph{Roundabout Merge}          & $> 4$                   & $25.20$                                        & $10.00$                                            & $21.00\pm 1.00$, $25.00\pm 1.00$, $35.00\pm 1.00$\footnotemark[3]                                                                                                                      & $0.00$                                 & $0.00$                                 \\ \hline
      \emph{Crossroad Turn Left}       & $> 7$                   & $25.20$                                        & $10.00$                                            & $26.00\pm 1.00$, $42.00\pm 1.00$\footnotemark[3]                                                                                                                                       & $0.00$                                 & $-42.00$                               \\ \hline
      \emph{Overtake}                  & $3$                     & $33.48$                                        & $11.00$                                            & $25.00\pm 1.00$\footnotemark[3] (Zombie Car \uppercase\expandafter{\romannumeral 1\relax}), $40.00\pm 1.00$\footnotemark[3] (Zombie Car \uppercase\expandafter{\romannumeral 2\relax}) & $0.00$                                 & $0.00$                                 \\
      \bottomrule[1pt]
    \end{tabular}
    \begin{tablenotes}
      \item \footnotemark[3]{The footnote part is the speed configuration for experimental results in Table~\ref{tab:quant-comparison} and Table~\ref{tab:policy-distillation-various-scenario}.
        % ; the results in Table \ref{tab:policy-distillation} are based on all speed configuration in Table \ref{tab:scenario-configuration}.
      }
    \end{tablenotes}
  \end{threeparttable}
\end{table*}

\paragraph{Data Collection} The demonstration data is collected by human driving ego car in an autonomous driving simulation platform. The simulation platform shown in Fig.\ref{fig:real} includes CARLA simulator with Logitech G29 Driving Force Steering Wheels $\&$ Pedals. The PC configuration is Ubuntu 16.04 x86, Intel Core i7-8700 CPU, GeForce GTX $1060$ and $16$ GB memory.
The following criteria are used to set to filter out low-quality demonstrations:
\begin{itemize}
  \item There is no collision;
        % \item Sufficient accumulated heuristic rewards in one episode;
  \item No dangerous moves and obeying traffic rules.
\end{itemize}
For each scenario, experts repeatedly drive in the simulator to finish episodes with $200$ steps. Trajectories violating the aforementioned criteria are rejected. After collecting $1000$ trajectories in each scenario, we each choose $100$ trajectories that meet the above criteria as a demonstration. To improve the generalization ability of the learned policy, stochasticity was added into the expert's steering and throttle during the collecting procedure. Namely, a random configuration is sampled at the beginning of each episode, the expert would choose different strategies to finish the current scenario. Aggressive strategies may include rash steering and throttle.

The statistic results of expert data are listed as Table~\ref{tab:expert_data}. Note that the ‘speed’ column in the Table~\ref{tab:expert_data} is not the expert data statistics but only used to distinguish sub-environments. For example $26.00\pm 1.00$ indicates the expert data collected under the \emph {Crossroad Turn Left} ($26.00\pm 1.00$) sub-environment. Statistical indicators include Collision Rate, Time Taken, Accel. and Jerk.

\begin{table*}[h]
  \centering
  \caption{Statistics of Expert Data}
  \label{tab:expert_data}
  % \begin{tabular}{cccccc}
  \begin{tabular}{>{\centering\arraybackslash}m{0.15\linewidth}>{\centering\arraybackslash}m{0.17\linewidth}>{\centering\arraybackslash}m{0.12\linewidth}>{\centering\arraybackslash}m{0.15\linewidth}>{\centering\arraybackslash}m{0.12\linewidth}>{\centering\arraybackslash}m{0.15\linewidth}}
    \toprule[1pt]
    \textbf{Scenarios}                                                 & \textbf{Speed ($km / h$)}        & \textbf{Collision Rate ($\%$)} & \textbf{Time Taken ($s$)} & \textbf{Accel. ($m/s^2$)} & \textbf{Jerk ($m/s^3$)} \\
    \toprule[1pt]
    \emph{Empty Town}                                                  & /                                & $0.00$                         & $280.21 \pm 10.48$        & $1.59 \pm 0.23$           & $4.25 \pm 0.21$         \\ \hline
    \emph{Single Lane Car Following}                                   & $15.00\pm 1.00$                  & $0.00$                         & $23.21 \pm 2.98$          & $2.01 \pm 0.59$           & $2.15 \pm 0.42$         \\ \hline
    \emph{Two Lanes Car Following}                                     & $18.00\pm 1.00$                  & $0.00$                         & $48.56 \pm 1.79$          & $2.62 \pm 0.51$           & $7.12 \pm 0.49$         \\ \hline
    \multirow{2}{1.5cm}[-0.05in]{\centering\emph{Crossroad Merge}}     & $21.00\pm 1.00$                  & $0.00$                         & $47.95 \pm 4.21$          & $2.42 \pm 0.45$           & $1.54 \pm 0.48$         \\ \cmidrule(lr){2-6}
                                                                       & $25.00\pm 1.00$                  & $0.00$                         & $48.87 \pm 1.21$          & $2.63 \pm 0.43$           & $1.69 \pm 0.19$         \\ \hline
    \multirow{3}{1.5cm}[-0.1in]{\centering\emph{Roundabout Merge}}     & $21.00\pm 1.00$                  & $0.00$                         & $36.23 \pm 3.69$          & $2.11 \pm 0.19$           & $1.69 \pm 0.60$         \\ \cmidrule(lr){2-6}
                                                                       & $25.00\pm 1.00$                  & $0.00$                         & $34.91 \pm 0.34$          & $2.01 \pm 0.32$           & $1.65 \pm 0.19$         \\ \cmidrule(lr){2-6}
                                                                       & $35.00\pm 1.00$                  & $0.00$                         & $42.17 \pm 0.59$          & $2.59 \pm 0.92$           & $1.69 \pm 0.21$         \\ \hline
    \multirow{2}{1.5cm}[-0.05in]{\centering\emph{Crossroad Turn Left}} & $26.00\pm 1.00$                  & $0.00$                         & $34.91 \pm 3.33$          & $1.93 \pm 0.21$           & $2.45 \pm 0.21$         \\
    \cmidrule(lr){2-6}
                                                                       & $42.00\pm 1.00$                  & $0.00$                         & $35.01 \pm 3.01$          & $2.19 \pm 0.11$           & $2.72 \pm 0.31$         \\ \hline
    \emph{Overtake}                                                    & $25.00\pm 1.00$, $40.00\pm 1.00$ & $0.00$                         & $33.52 \pm 1.65$          & $4.99 \pm 0.51$           & $4.95 \pm 0.55$         \\
    \bottomrule[1pt]
  \end{tabular}
\end{table*}

\paragraph{Evaluation Metrics} We compare the learned decision module with both the rule-based module and the end-to-end neural control policy. Rule-based module is composed following \cite{shalev2017formal}. The end-to-end policy is trained with vanilla GAIL. Following \cite{bansal2019combining}, we conduct quantitative comparison including metrics like collision rate, time to accomplish tasks, average acceleration, and jerk after $100$ trials in evaluation environments. These metrics reflect how safe and smooth the agent drives. The trajectory from different approaches is also visualized and compared qualitatively. We provide demo videos in the supplementary.

\subsection{Baseline Methods}
\paragraph{Rule-based Method} The rule-based system in our paper is based on the formal model provided in the paper of MobileEye~\cite{shalev2017formal}. This paper proposed the definitions of the safe longitudinal and lateral distance and designed a car-following model. For example, the vehicle should keep the safe longitudinal distance to the front car and avoid collision with the rear car. Since the paper didn’t mention the overtake scenario, we divided the overtake scenario into three stages and design rules with the instructions of the formal model. In the first stage, the car-following stage, the vehicle would make decisions instructed by the car-following model. In the second stage, the turning-left stage, the agent would decide when to turn left according to the safe longitudinal and lateral distance, then turning-left with high speed. In the last stage, the turning-right stage, similar to the turning-left decision, the agent would turn right back at a reasonable speed.
\paragraph{End-to-end Neural Policy} The end-to-end method used in our paper is GAIL~\cite{ho2016generative}. Nothing has changed except that the state space and action space have been adapted to the CARLA environment.

\subsection{Other Training Details}
\paragraph{Reparameterization} The planning and control module are both deterministic. Hence at every state, once the decision is chosen, the transformation from decision to control is a deterministic and correspondence mapping. The reparameterization trick can be seen in Equation~\ref{eq:reparam}. We reparametrize $u$ in the first expectation with $d$. In other words, the conditional probability of $d$ is the same as $u$.

\paragraph{Decision Space} Human decisions can be divided into three parts: which lane to go, how far to go, at which speed to go. We model the three parts as three categorical variables to simplify its representations. To make a tractable inversion of the collected data, a 10 meters interval is used in the longitudinal goal while a 10km/h interval in the target speed. The interval is based on the feedback from our human drivers.
\paragraph{Policy Architecture and Hyperparameters} % Learning-based Module
The policy architecture of the decision module in our method is exactly the same as the architecture in the paper of GAIL~\cite{ho2016generative}. For a fair comparison, all hyperparameters are also the same. Learning-based Module method and End-to-end Neural Policy baseline are trained for 500 iterations before evaluation while Rule-based method baseline does not require training and can be evaluated directly. The maximum time step for each iteration is $1000$. We use a two-layer ReLU network with 32 units for the discriminator of GAIL and our method. For the policy, we use a two-layer (32 units) ReLU neural network. Entropy regularizer weight is set to be 0.1 across our method and GAIL. We use a batch size of $512$ steps per update. The learning rate for generator and discriminator is $0.0003$.

\section{More Experiment Results}
\subsection{Policy Distillation (Generalization Capability)}
The disadvantage of end-to-end learning is not just that there are no guarantees, but also that generalization is tricky. Generalization capability is crucial for our method. There are three experiments that provide evidence for the generalization capability of our method.
\begin{itemize}
  \item Different towns: The Empty town model trained in Town3 can be applied in Town2 and Town1 (cf. attached video).
  \item Different scenarios: The model trained in \emph{Crossroad Merge} also performs well in \emph{Roundabout Merge} and \emph{Single Lane Car Following} (cf. \emph{One Policy for Various Scenarios}).
  \item Different velocities: The \emph{Crossroad Merge} model trained with the velocity of 25km/h also performs well at 15km/h (cf. \emph{One Policy for Various Speeds}).
\end{itemize}

The aforementioned experiments train one specific model for each scenario separately. To test the practical potential of the proposed framework, we train one policy with demonstration data for multiple scenarios and multiple speeds. This policy distillation problem is divided into:  \uppercase\expandafter{\romannumeral 1\relax}) \emph{One Policy for Various Scenarios} and \uppercase\expandafter{\romannumeral 2\relax}) \emph{One Policy for Various Speeds}.
% As shown in Table \ref{tab:policy-distillation}, it achieves similar performance as above, successfully distilling human decision policy from all of them.  %\bz{I think we should include more description in this paragraph, and include more qualitative and quantitative results, as we consider this as one of the main contributions.}

For \emph{One Policy for Various Scenarios}, the policy distillation process runs as follows: We leverage the same demonstration data as the previous experiments in four scenarios (\emph{Single Lane Car Following, Crossroad Merge, Roundabout Merge} and \emph{Crossroad Turn Left}) together to train one policy for these scenarios jointly. Four CARLA simulators run in parallel, each corresponding to one scenario, while only one learner is deployed to learn a policy from all data sampled simultaneously. Then we can obtain one policy model for all scenarios.

For \emph{One Policy for Various Speeds}, we leverage the same demonstration data for each scenario. However, the demonstration data for each scenario contains various speeds (as Table \ref{tab:scenario-configuration} shows). We train a policy for each scenario using corresponding demonstration data and environments at various speeds, while only one learner is deployed to learn one policy from all data sampled from different speeds simultaneously. Finally, one policy is learned for various speeds.

Table \ref{tab:policy-distillation-various-scenario} shows the statistics of behaviors of the learning-based modular system for various scenarios. After the training is finished, the learned policy is evaluated on the evaluation environment of different scenarios respectively (100 trials each). We also evaluated the learned policy on the other two unseen environments and reported the average results of 100 trials in Table~\ref{tab:policy-distillation-various-scenario}. Table~\ref{tab:policy-distillation-various_speeds-crossroad-merge}~\ref{tab:policy-distillation-various_speeds-roundabout-merge}~\ref{tab:policy-distillation-various_speeds-crossroad-turn-left} shows the statistics of behaviors of the learning-based modular system for various speeds. The learned policy is evaluated on the evaluation environment of various speeds respectively (100 trials each). The distilled policy shows a slight performance loss compared to the policies previously trained separately in the case of one specific scenario. But the performance degradation is within the acceptable range. %The performance of distilled policy is reasonable with regard to collision rate,  time taken,  acceleration, and jerk. 
The above policy distillation results (\emph{One Policy for Various Scenarios} and \emph{One Policy for Various Speeds}) exhibits the potential that our framework can handle scenarios in different levels of complexity concurrently without compromising performance.

\begin{table}[htp!]
  \centering
  \caption{Generalized to Various Scenarios}
  \label{tab:policy-distillation-various-scenario}
  \begin{tabular}{>{\centering\arraybackslash}m{0.28\linewidth}>{\centering\arraybackslash}m{0.12\linewidth}>{\centering\arraybackslash}m{0.15\linewidth}>{\centering\arraybackslash}m{0.12\linewidth}>{\centering\arraybackslash}m{0.12\linewidth}}
    \toprule[1pt]
    \multirow{2}{*}[-0.15in]{\textbf{Scenarios}}
    % & \multirow{2}{2cm}[-0.15in]{\textbf{Speed \\ ($km / h$)}}   
                                           & \multicolumn{4}{c}{\textbf{Learning-based Module}}                                                                                   \\
    \cmidrule(lr){2-5}
                                           & \textbf{Collision Rate ($\%$)}                     & \textbf{Time Taken ($s$)} & \textbf{Accel. ($m/s^2$)} & \textbf{Jerk ($m/s^3$)} \\
    \toprule[1pt]
    \emph{Single Lane Car Following}       & $2.00$                                             & $24.35 \pm 1.01$          & $2.41 \pm 0.56$           & $1.92 \pm 0.22$         \\
    \hline
    % \multirow{2}{1cm}[-0.05in]{\centering\emph{Crossroad \\Merge}}
    \emph{Crossroad  Merge}                & $8.00$                                             & $52.83 \pm 0.61$          & $2.92 \pm 0.21$           & $5.31 \pm 0.33$         \\ %\cmidrule(lr){2-6}
    \hline
    % \multirow{3}{1cm}[-0.1in]{\centering\emph{Roundabout \\Merge}}
    \emph{Roundabout Merge}                & $14.00$                                            & $40.99 \pm 1.26$          & $2.21 \pm 0.67$           & $1.82 \pm 0.32$         \\ %\cmidrule(lr){2-6}
    \hline
    %   & 25.00         & 0.00              & 39.56 & 1.99         & 1.70   \\ \cmidrule(lr){2-6}
    % & 35.00         & 0.00              & 37.15 & 2.01         & 1.72    \\ \hline
    % \multirow{2}{1.5cm}[-0.05in]{\centering\emph{Crossroad \\Turn Left}} 
    \emph{Crossroad Turn Left}             & $14.00$                                            & $35.21 \pm 2.29$          & $4.81 \pm 0.95$           & $5.22 \pm 0.68$         \\ \hline
    % \cmidrule(lr){2-6}
    %   & 42.00    & 0.00              & 36.53 & 2.01         & 2.72   \\ 
    \emph{Two Lane Car Following} (Unseen) & $6.00$                                             & $50.35 \pm 4.81$          & $2.81 \pm 0.61$           & $7.92 \pm 0.53$         \\
    \hline
    % \multirow{2}{1cm}[-0.05in]{\centering\emph{Crossroad \\Merge}}
    \emph{Overtake}      (Unseen)          & $18.00$                                            & $35.62 \pm 1.16$          & $5.63 \pm 0.55$           & $5.43 \pm 0.72$         \\ %\cmidrule(lr){2-6}
    \bottomrule[1pt]
  \end{tabular}
\end{table}

\begin{table}[htp!]
  \centering
  \caption{Generalized to Various Speeds (Crossroad Merge)}
  \label{tab:policy-distillation-various_speeds-crossroad-merge}
  \begin{tabular}{>{\centering\arraybackslash}m{0.12\linewidth}>{\centering\arraybackslash}m{0.125\linewidth}>{\centering\arraybackslash}m{0.12\linewidth}>{\centering\arraybackslash}m{0.15\linewidth}>{\centering\arraybackslash}m{0.12\linewidth}>{\centering\arraybackslash}m{0.12\linewidth}}
    \toprule[1pt]
    \multirow{2}{*}[-0.15in]{\textbf{Scenarios}} & \multirow{2}{2cm}[-0.15in]{\textbf{Speed                                                                                                                    \\ ($km / h$)}}   & \multicolumn{4}{c}{\textbf{Learning-based Module}} \\
    \cmidrule(lr){3-6}
                                                 &                                          & \textbf{Collision Rate ($\%$)} & \textbf{Time Taken ($s$)} & \textbf{Accel. ($m/s^2$)} & \textbf{Jerk ($m/s^3$)} \\
    \toprule[1pt]
    \multirow{2}{1.5cm}[-0.05in]{\centering\emph{Crossroad                                                                                                                                                     \\Merge}} &  $21.00 \pm 1.00$          & $10.00$              & $51.82 \pm 4.32$ & $2.99 \pm 0.44$        & $5.52 \pm 0.79$  \\ \cmidrule(lr){2-6}
                                                 & $25.00  \pm 1.00$                        & $12.00$                        & $47.83 \pm 5.21$          & $2.83 \pm 0.35$           & $5.69 \pm 0.58$         \\
    %   \hline
    % \multirow{3}{1cm}[-0.1in]{\centering\emph{Roundabout \\Merge}}  &  21.00         & 0.00              & 40.99 & 2.21         & 1.82   \\ \cmidrule(lr){2-6}
    %   & 25.00         & 0.00              & 39.56 & 1.99         & 1.70   \\ \cmidrule(lr){2-6}
    % & 35.00         & 0.00              & 37.15 & 2.01         & 1.72    \\ \hline
    % \multirow{2}{1.5cm}[-0.05in]{\centering\emph{Crossroad \\Turn Left}}  & 26.00     & 0.00              & 42.55 & 1.95         & 2.51   \\ 
    % \cmidrule(lr){2-6}
    %   & 42.00    & 0.00              & 36.53 & 2.01         & 2.72   \\ 
    \bottomrule[1pt]
  \end{tabular}
\end{table}

\begin{table}[htp!]
  \centering
  \caption{Generalized to Various Speeds (Roundabout Merge)}
  \label{tab:policy-distillation-various_speeds-roundabout-merge}
  \begin{tabular}{>{\centering\arraybackslash}m{0.13\linewidth}>{\centering\arraybackslash}m{0.125\linewidth}>{\centering\arraybackslash}m{0.12\linewidth}>{\centering\arraybackslash}m{0.15\linewidth}>{\centering\arraybackslash}m{0.12\linewidth}>{\centering\arraybackslash}m{0.12\linewidth}}
    \toprule[1pt]
    \multirow{2}{*}[-0.15in]{\textbf{Scenarios}} & \multirow{2}{2cm}[-0.15in]{\textbf{Speed                                                                                                                    \\ ($km / h$)}}   & \multicolumn{4}{c}{\textbf{Learning-based Module}} \\
    \cmidrule(lr){3-6}
                                                 &                                          & \textbf{Collision Rate ($\%$)} & \textbf{Time Taken ($s$)} & \textbf{Accel. ($m/s^2$)} & \textbf{Jerk ($m/s^3$)} \\
    \toprule[1pt]
    % \multirow{2}{1cm}[-0.05in]{\centering\emph{Crossroad \\Merge}} &  21.00          & 0.00              & 51.83 & 2.92         & 5.31   \\ \cmidrule(lr){2-6}
    %   &   25.00        & 0.00              & 47.83 & 2.83         & 5.29    \\ \hline
    \multirow{3}{1.5cm}[-0.1in]{\centering\emph{Roundabout                                                                                                                                                     \\Merge}}  &  $21.00 \pm 1.00$         & $14.00$              & $41.52\pm 5.22$ & $2.63 \pm 0.31$         & $1.72 \pm 0.35$  \\ \cmidrule(lr){2-6}
                                                 & $25.00 \pm 1.00$                         & $12.00$                        & $39.56 \pm 6.21$          & $2.09 \pm 0.39$           & $1.70 \pm 0.41$         \\ \cmidrule(lr){2-6}
                                                 & $35.00 \pm 1.00$ (Unseen)                & $14.00$                        & $37.15 \pm 5.59$          & $2.01 \pm 0.49$           & $1.72 \pm 0.42$         \\
    % \hline
    % \multirow{2}{1.5cm}[-0.05in]{\centering\emph{Crossroad \\Turn Left}}  & 26.00     & 0.00              & 42.55 & 1.95         & 2.51   \\ 
    % \cmidrule(lr){2-6}
    %   & 42.00    & 0.00              & 36.53 & 2.01         & 2.72   \\ 
    \bottomrule[1pt]
  \end{tabular}
\end{table}

\begin{table}[htp!]
  \centering
  \caption{Generalized to Various Speeds (Crossroad Turn Left)}
  \label{tab:policy-distillation-various_speeds-crossroad-turn-left}
  \begin{tabular}{>{\centering\arraybackslash}m{0.15\linewidth}>{\centering\arraybackslash}m{0.125\linewidth}>{\centering\arraybackslash}m{0.12\linewidth}>{\centering\arraybackslash}m{0.15\linewidth}>{\centering\arraybackslash}m{0.12\linewidth}>{\centering\arraybackslash}m{0.12\linewidth}}
    \toprule[1pt]
    \multirow{2}{*}[-0.15in]{\textbf{Scenarios}} & \multirow{2}{2cm}[-0.15in]{\textbf{Speed                                                                                                                    \\ ($km / h$)}}   & \multicolumn{4}{c}{\textbf{Learning-based Module}} \\
    \cmidrule(lr){3-6}
                                                 &                                          & \textbf{Collision Rate ($\%$)} & \textbf{Time Taken ($s$)} & \textbf{Accel. ($m/s^2$)} & \textbf{Jerk ($m/s^3$)} \\
    \toprule[1pt]
    % \multirow{2}{1cm}[-0.05in]{\centering\emph{Crossroad \\Merge}} &  21.00          & 0.00              & 51.83 & 2.92         & 5.31   \\ \cmidrule(lr){2-6}
    %   &   25.00        & 0.00              & 47.83 & 2.83         & 5.29    \\ \hline
    % \multirow{3}{1cm}[-0.1in]{\centering\emph{Roundabout \\Merge}}  &  21.00         & 0.00              & 40.99 & 2.21         & 1.82   \\ \cmidrule(lr){2-6}
    %   & 25.00         & 0.00              & 39.56 & 1.99         & 1.70   \\ \cmidrule(lr){2-6}
    % & 35.00         & 0.00              & 37.15 & 2.01         & 1.72    \\ \hline
    \multirow{2}{1.5cm}{\centering\emph{Crossroad                                                                                                                                                              \\Turn Left}}  & $26.00 \pm 1.00$     & $14.00$              & $36.49 \pm 3.21$ & $2.15 \pm 0.29$         & $2.99 \pm 0.32$   \\
    \cmidrule(lr){2-6}
                                                 & $42.00 \pm 1.00$                         & $16.00$                        & $38.53 \pm 3.18$          & $2.09 \pm 0.81$           & $2.72 \pm 0.69$         \\
    \bottomrule[1pt]
  \end{tabular}
\end{table}
\end{document}

\maketitle

%===============================================================================

\appendix
\section*{Appendices}
\addcontentsline{toc}{section}{Appendices}
\renewcommand{\thesection}{\Alph{section}}
\label{sec:appendix}
The appendices are organized as follows: In Section \ref{sec:theoretical}, a general theoretical understanding of IABP is presented. In Section \ref{sec:energy_matching}, we distinguish our method from Contrastive Divergence (CD) theoretically. In Section \ref{sec:exp_details}, we introduce the experimental details respectively, including dataset splitting in Section \ref{sec:dataset}, network structure in Section \ref{sec:network}, hyper-parameters for different tasks in Section \ref{sec:recon}, the impact of Langevin steps in Section \ref{sec:appendix_lang_step}, latent adversarial attack and a robustness comparison with existing methods in Section \ref{sec:appendix_attack}. In Section \ref{sec:appendix_disentanglement}, we show more results of unsupervised  hierarchical latent disentanglement to demonstrate that HiABP can not only well recover the pixel values of the input image, but also align the image with the rich semantics encoded in latent space. 
\section{Policy Architecture and Hyper-parameters}
The policy architecture of decision module is exactly the same architecture in the paper of GAIL. Other hyperparameters are the same too.

%===============================================================================

% The maximum paper length is 8 pages excluding references and acknowledgements, and 10 pages including references and acknowledgements

\clearpage
% The acknowledgments are automatically included only in the final version of the paper.
\acknowledgments{If a paper is accepted, the final camera-ready version will (and probably should) include acknowledgments. All acknowledgments go at the end of the paper, including thanks to reviewers who gave useful comments, to colleagues who contributed to the ideas, and to funding agencies and corporate sponsors that provided financial support.}

%===============================================================================

\bibliography{example}  % .bib